\documentclass{article}

    \PassOptionsToPackage{numbers, compress}{natbib}

\usepackage[main,final]{neurips_2026}

\usepackage[utf8]{inputenc}
\usepackage[T1]{fontenc}
\usepackage{hyperref}
\hypersetup{hidelinks}
\usepackage{url}
\usepackage{booktabs}
\usepackage{amsfonts}
\usepackage{nicefrac}
\usepackage{microtype}
\usepackage[table]{xcolor}
\usepackage{graphicx}
\usepackage{subcaption}
\usepackage{amsmath}
\usepackage{amssymb}
\usepackage{mathtools}
\usepackage{amsthm}
\usepackage{booktabs}
\usepackage{cellspace}
\usepackage{longtable}
\usepackage{array}
\usepackage{tabularx}
\RequirePackage{algorithm}
\RequirePackage{algorithmic}

\usepackage[disable,textsize=tiny]{todonotes}

\title{SEAR: Sample Efficient Action Chunking Reinforcement Learning}

\author{%
  C F Maximilian Nagy\textsuperscript{1,2} \And
  Onur Celik\textsuperscript{1} \And
  Emiliyan Gospodinov\textsuperscript{1} \And
  Florian Seligmann\textsuperscript{1} \And
  Weiran Liao\textsuperscript{1} \And
  Aryan Kaushik\textsuperscript{1} \And
  Gerhard Neumann\textsuperscript{1,2} \\\\
  \textsuperscript{1}Autonomous Learning Robots, Karlsruhe Institute of Technology \\
  \textsuperscript{2}FZI Forschungszentrum Informatik, Karlsruhe \\
  \texttt{nagy@fzi.de}
}

\begin{document}

\maketitle

\begin{abstract}
Action chunking improves exploration and accelerates value propagation in long-horizon reinforcement learning, but naively applying off-policy methods to the temporally extended action space at reduced decision frequency offsets these gains, leading to poor sample efficiency. Existing online action chunking methods address these issues through computationally expensive critic-only approaches or by relying on offline data. We introduce SEAR, a sample-efficient off-policy algorithm that enables online reinforcement learning with action chunks by addressing both challenges. To handle high-dimensional action sequences, SEAR employs a causal transformer critic trained with multi-horizon targets that provide a training signal for every prefix of an action chunk, effectively increasing the useful gradients per sample. To restore decision frequency, SEAR operates with a receding horizon and random replanning, ensuring uniform state coverage while combining the fast value propagation of large chunks with the reactivity of small ones. SEAR outperforms state-of-the-art online reinforcement learning methods including SimbaV2 on challenging Metaworld manipulation tasks. Beyond online RL, applying SEAR to the offline-to-online method QC improves its performance on the demanding OGBench cube-triple tasks.
\end{abstract}
\begin{center}
\url{https://alrhub.github.io/SEAR/}
\end{center}

\section{Introduction}

Action chunking has seen remarkable success in imitation learning (IL)~\cite{chi2024diffusionpolicy,zhao2023act,reuss2025efficient,black2410pi0,intelligence2025pi06vlalearnsexperience,
li2025grrlgoingdexterousprecise,
zheng2025xvlasoftpromptedtransformerscalable,
nikolov2025spear1scalingrobotdemonstrations,
lee2025molmoactactionreasoningmodels}.
There, a policy is trained to predict a sequence of future actions instead of a single action, which are then executed in an open-loop manner, enabling the policy to capture coherent multi-step behaviors.%

Building on these results, a growing line of work explores action chunking reinforcement learning (RL)~\cite{li2024top, li2025qchunk, li2025decoupledqchunking, yang2025ac3, park2025scalableofflinemodelbasedrl, seo2024coarse}.
For reinforcement learning, action chunking enables faster and more stable reward propagation through unbiased N step returns computed over executed chunks, which is particularly beneficial in long horizon and sparse reward settings~\cite{li2025qchunk,li2025decoupledqchunking}. By operating over temporally extended action sequences, action chunking facilitates the discovery of meaningful skills and improves exploration by sampling coherent action trajectories rather than relying on single time step random walk noise. This structured exploration leads to more consistent state space coverage and can substantially reduce exploration burden. Additionally, action chunking provides a natural interface for incorporating offline data when available, allowing learned temporal patterns to be reused effectively during online learning~\cite{li2025qchunk}.

However, action chunking significantly increases the action space of the policy, requiring the critic to evaluate chunks rather than individual actions.
In this enlarged action space, critic approximation errors can lead to poor actor updates~\cite{seo2024coarse}, requiring careful algorithm design.
In particular, naively applying action chunking to off-the-shelf single-action RL algorithms significantly reduces performance, as shown in Figure~\ref{fig:comparison}a.

Moreover, executing $N$ actions in an open-loop manner reduces the policy’s decision frequency, as 
the policy is only queried every $N$ steps. This has two consequences: Firstly, it requires the predicted action sequence to be highly 
accurate, as errors cannot be corrected during execution~\cite{li2025decoupledqchunking}. 
Secondly, this reduces the policy's ``state-action coverage'', especially if the initial state 
distribution is narrow (Figure~\ref{fig:random_init}).

A common way to restore policy reactivity is receding horizon control, where the policy predicts an 
action chunk of length $N$ but executes only a prefix of $k$ actions before replanning. While this 
strategy is widely used in imitation learning to balance long horizon planning with closed-loop 
control, it has so far not been explored for action chunking policies in model-free
reinforcement learning. However,
applying receding horizon control in reinforcement learning is nontrivial, as executing shorter 
prefixes at evaluation time introduces a distribution shift from training on fixed length action 
chunks.

We introduce \textit{Sample Efficient Action Chunking Reinforcement Learning (SEAR)}, an off-policy RL algorithm that learns action chunking policies. SEAR uses a \textit{multi-horizon} critic target that captures the causal relation between a chunk's
actions while providing per-action training targets. Combined with a causal transformer critic,
this enables stable learning even for large chunk sizes and even when no prior data is available. Furthermore, SEAR
improves the state-action coverage of the actor using \textit{random replanning}, where only a random prefix of each chunk is executed during data collection. Together, random replanning and multi-horizon targets enable SEAR to effectively use receding horizon control to decouple the long action chunks useful during training with shorter evaluation action chunk sizes.
In the extreme case, SEAR can also use action chunking only during training, leading to better single-step success rates than state-of-the-art non-chunked methods.

\begin{figure*}
    \centering
    \begin{subfigure}[b]{.57\textwidth}
        \centering
        \includegraphics[width=\linewidth]{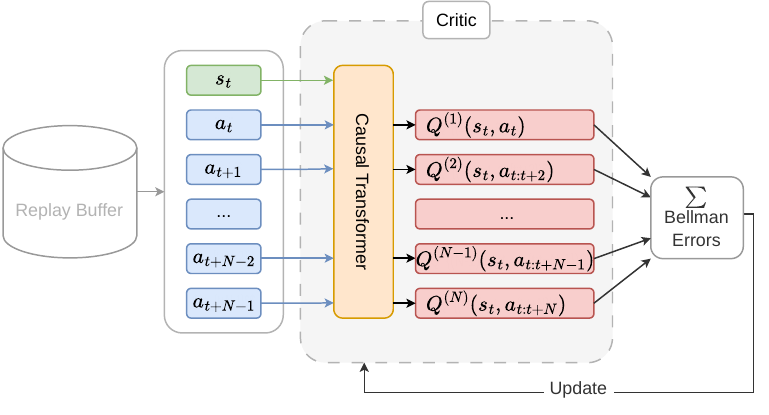}
        \caption{Critic Update}
        \label{fig:critic-design}
    \end{subfigure}
    \hfill
    \begin{subfigure}[b]{.42\textwidth}
        \centering
        \includegraphics[width=.98\linewidth]{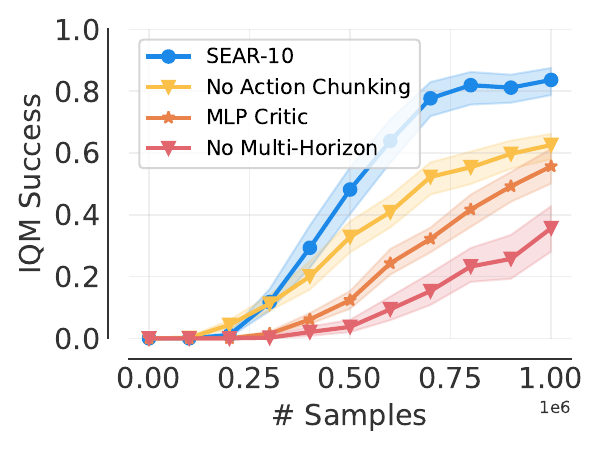}
        \caption{Critic Ablations (Metaworld-Hard)}
        \label{fig:replanning_ablation}
    \end{subfigure}
    \caption{\textbf{Overview of SEAR's (a) critic design and (b) how it affects performance.}
    The critic function expects the state $s_t$ and an action chunk
    $a_{t:t + N}$ of chunk size $N$ as argument. During the critic
    update \textbf{(a)}, these training points are sampled from the
    replay buffer and the causal transformer critic predicts Q-values
    $Q^{(1)}(s_t,a_t),Q^{(2)}(s_t,a_{t:t+2}),\dots,Q^{(N)}(s_t,a_{t:t+N})$
    for all action chunk prefixes, thereby generating
    \textit{multi-horizon} predictions that increase the amount
    of training data for the subsequent critic update. Only fitting the
    critic on the whole-chunk target $Q^{(N)}$, as is common in offline
    RL~\cite{li2025qchunk,zhang2026sacflowsampleefficientreinforcement},
    does not work well in online RL \textbf{(b)}. Furthermore, bootstrapping an action
    chunking critic with a single-step policy, comparable to T-SAC (``No
    Action Chunking'') lowers success, as does replacing the transformer
    backbone with an MLP.}
    \label{fig:critic}
\end{figure*}

We evaluate SEAR on the challenging Metaworld~\cite{yu2021metaworldbenchmarkevaluationmultitask,mclean2026metaworld} online RL benchmark and the OGBench~\cite{ogbench_park2025} offline-to-online RL benchmark.
On Metaworld, SEAR achieves a substantial improvement of over 20 percentage points in IQM success rate~\cite{agarwal2021deep} compared to previous methods among the 20 hardest environments, as shown in Figure~\ref{fig:comparison}. Through extensive ablation studies, we highlight the impact of each design decision of our approach, such as using a causal transformer critic with and without multi-horizon targets, and random replanning (Figure~\ref{fig:critic}, Figure~\ref{fig:random_init}). We further evaluate the impact of different chunk sizes (Figure~\ref{fig:ablations-chunksize}). In particular, we demonstrate that receding horizon control, which our work is the first to apply to action chunking policies in model-free RL, is only effective if the main components of our algorithms are used, i.e., multi-horizon targets, causal transformer critics, and random replanning during training.  

We summarize our contributions as follows: 
\begin{itemize}
\item We introduce SEAR, an off-policy RL method for action chunking grounded in the maximum-entropy RL framework. 
\item We incorporate a causal transformer critic and multi-horizon N-step returns, leading to improved sample efficiency.
\item We increase the state-action coverage by executing only a random prefix of each action chunk before replanning, complementing a receding horizon execution mode for fast reactivity.
\item Through extensive experiments, we demonstrate that SEAR consistently improves the performance for online and offline-to-online RL, with an IQM success rate over 20 percentage points higher on the 20 hardest Metaworld environments. In ablations, we show the importance and performance impact of all design decisions of our action chunking approach. 
\end{itemize}

\section{Related Work}

In robotic learning, action chunking agents generate
multiple consecutive actions at once. This was
first introduced in the context of imitation learning
by~\citet{zhao2023act} and was widely adopted by other imitation learning
approaches~\citep{chi2024diffusionpolicy,zhao2023act,reuss2025efficient,black2410pi0,intelligence2025pi06vlalearnsexperience,
li2025grrlgoingdexterousprecise,
zheng2025xvlasoftpromptedtransformerscalable,
nikolov2025spear1scalingrobotdemonstrations,
lee2025molmoactactionreasoningmodels}.

\paragraph{N-step Returns in Reinforcement Learning}
In reinforcement learning, N-step returns have a long history as a
compromise between unbiased, high-variance Monte-Carlo returns and
high-bias, low-variance TD returns~\citep{sutton1998reinforcement}. A
fundamental problem of applying N-step returns to off-policy algorithms
is that the action-, and therefore also the reward-distribution, of the replay
buffer differs from that of the current policy. The standard way to mitigate
this issue is importance sampling, where care must be taken to reduce
variance~\cite{munos2016retracelambda}.

A principled way to avoid this bias is to use an action-chunk-conditioned
critic where the N-step return is computed over the given chunk. This was first proposed
in the TOP-ERL~\cite{li2024top} algorithm for the episodic reinforcement
learning setting and has since been adapted to traditional single-action
policies~\cite{dong2025chunkingthecritic} and action chunking
policies~\cite{li2025qchunk,li2025decoupledqchunking,yang2025ac3,park2025scalableofflinemodelbasedrl,seo2024coarse}.

\paragraph{Multi-Target Updates for Action Chunking Critics}

A fundamental aspect of learning Q-functions with action chunking is the increased dimensionality of the resulting action space, rendering Q-function approximation more difficult. 
Most action chunking methods use critics that map from this extended action space to a single Q-target~\cite{li2025qchunk,li2025decoupledqchunking,yang2025ac3,park2025scalableofflinemodelbasedrl}.
For these methods, increasing the chunk size $N$ increases the action space dimension without introducing new 
training targets. Experimental evidence suggests that these critic fitting approaches are unsuitable for online RL (cf Section~\ref{sec:ablations}, \cite{seo2024coarse}). However, in the offline and offline-to-online RL setting, where
demonstration data is available as a prior for learning, this approach can be used effectively~\cite{li2025qchunk,li2025decoupledqchunking,yang2025ac3,park2025scalableofflinemodelbasedrl}.

When the critic instead learns a Q-target for each action of a chunk, the ratio between action space dimension and number of training targets becomes independent of the chunk size $N$. 
This kind of multi-horizon critic was first proposed in the episodic
RL setting by TOP-ERL~\cite{li2024top}. TOP-ERL uses a trajectory generator to generate actions for the whole episode at
once. Segments of these trajectories are then used for off-policy updates using the reparametrization trick~\cite{haarnoja2018soft}. For each prefix of these segments, an N-step Q-target is calculated and used for critic updates. In contrast to the action chunking setting, the segment length in TOP-ERL can be varied without affecting the action space dimension.
T-SAC~\cite{dong2025chunkingthecritic} adapts TOP-ERL's critic for standard policies without action chunking.

CQN-AS~\cite{seo2024coarse} is the only previous
action-chunking method that uses a multi-target critic. Its critic is trained to predict the returns of each
individual action instead of the N-step return of the combined actions. 
SEAR is the first method that fully exploits the temporal structure of action chunks by adapting TOP-ERL's
critic structure into the action chunking setting. This allows SEAR to be used for online RL, in contrast to methods
that treat each chunk as a single data point. These methods are typically confined to the offline-to-online RL case and do not work well for online RL. 
In contrast to CQN-AS, SEAR explicitly models the causal dependencies between actions.

\paragraph{Exploration in RL}
Balancing exploration and exploitation is a fundamental challenge in continuous reinforcement learning.
Some algorithms employ state-independent, fixed exploration noise on top of an otherwise deterministic
policy~\cite{lillicrap2015ddpg,fujimoto2018td3,yang2025ac3,seo2024coarse}. A common way to introduce state-dependent
exploration are Maximum-Entropy approaches, that learn when and how to explore~\cite{haarnoja2018soft,ball2023rlpd,celik2025dime}. Another approach to 
state-dependent exploration is to use an expressive imitation learning policy to guide 
exploration~\cite{li2025qchunk}.
SEAR is the first online RL algorithm that learns an action chunking policy in the maximum entropy setting.

\section{Preliminaries}\label{sec::preliminaries}

\textbf{Notation.} We aim to train a reinforcement learning policy $\pi : \mathcal{S} \times \mathcal{A}\rightarrow \mathbb{R}^+$ in the continuous state-action space $\mathcal{S} \times \mathcal{A}$ using the Markov decision process (MDP) formalization. 
The MDP is defined by the tuple $(\mathcal{S}, \mathcal{A}, P, r, \gamma, \rho_0)$, where $\gamma \in [0,1) $ is the discount factor, $P: \mathcal{S} \times \mathcal{A} \times \mathcal{S} \rightarrow \mathbb{R}^+$ is the transition probability function returning the likelihood for transitioning to state $s_{t+1} \in \mathcal{S} $ when executing action $a_t \in \mathcal{A}$ in state $s_t \in \mathcal{S}$ and $r: \mathcal{S} \times \mathcal{A} \rightarrow [r_{\min}, r_{\max}]$ is the bounded reward function. 
The initial state distribution $\rho_0: \mathcal{S} \rightarrow \mathbb{R}^+$ denotes the probability of starting at state $s_0\in\mathcal{S}$. 
Similar to \citet{haarnoja2018soft}, we denote $\rho^\pi$ as the state or state-action distribution under the current policy $\pi$
and denote the time step of a (random) variable with a subscript. 
This subscript applies to the reward in the sense that $r_t \triangleq r(s_t, a_t)$, which allows for a more compact notation.

We extend the single-step setup to a multi-step action chunking setup, where we overload the subscript notation when referring to a chunksize of $N$ by writing $a_{t:t+N}$, which denotes the sequence of $N$ actions $(a_t, a_{t+1}, ..., a_{t+N-1})$. 
Additionally, we refer to $\rho$ as the state-action distribution induced by the action chunking policy  $\pi(a_{t:t+N}|s_t)$ and which provides the density of a sequence of state-action chunks. 
Naturally, the dimensionality of the chunk-extended action space grows linearly with chunk size $N$ following $\dim(a_{t:t+N}) = N\dim(a_t)$.%

\textbf{Maximum Entropy Reinforcement Learning.} 
We consider the maximum entropy reinforcement learning (MaxEnt-RL) setup that augments the reward function with the entropy of the current policy, leading to the objective 
\begin{align}\label{eq:max_ent_obj}
    J(\pi) = \sum_{t=0}^\infty\gamma^{t} \mathbb{E}_{\rho^\pi}\left[r_t +\alpha \mathcal{H}(\pi(a_t|s_t))\right].
\end{align}
Here, $\alpha \in \mathbb{R}^+$ is a scaling factor that trade-offs maximizing the entropy $\mathcal{H}(\pi(a|s))=-\int \pi(a|s)\log\pi(a|s)$ against maximizing the task return and thereby plays a crucial role in exploration control. 
This objective also leads to an entropy augmented definition of the Q-function under the current policy $\pi$
\begin{align}\label{eq::MaxEnt_Q}
    Q^\pi(s_t,a_t) = r_t + \sum_{i=1}^\infty \gamma^i\mathbb{E}_{\rho^\pi}\left[r_{t+i} + \alpha \mathcal{H}(\pi(a_{t+i}|s_{t+i})) \right],
\end{align} with $Q^\pi: \mathcal S \times \mathcal A \mapsto \mathbb R$.
The MaxEnt-RL objective \cite{ziebart2008maximum, ziebart2010modeling,haarnoja2017reinforcement} has several benefits over the standard RL objective \cite{sutton1998reinforcement}, among which improved exploration is the key reason we base our method on it. 
\section{Sample Efficient Action Chunking RL}
\label{sec:method}
Naively applying action chunking to a policy does not yield satisfying results (see Figure~\ref{fig:comparison} and \citet{seo2024coarse}). This section introduces SEAR's elementary algorithmic and design choices that are required to obtain action-chunking policies that are sample efficient and achieve high success rates. 
We start by formalizing the MaxEnt RL objective for action chunking policies, which we use to define the action-chunk-conditioned Q-function.
Based on this definition, we introduce multi-horizon Q-function targets that effectively increase the number of target points for training the respective Q-function. This Q-function allows for straightforward policy updates, which we present in the subsequent section. Finally, we propose random replanning that plays a significant role in state coverage and receding horizon replanning. Figure~\ref{fig:critic} summarizes the update procedure for the critic. Algorithm~\ref{alg:sear} in Appendix~\ref{app:algorithm} summarizes the full training procedure.

\subsection{Action Chunking Maximum Entropy RL}
We aim to optimize an action chunking policy $\pi(a_{t:t+N} \mid s_t)$ that predicts an action sequence $a_{t:t+N}$ of length $N$ given the current state $s_{t}$.
For this, we start by formalizing the infinite horizon MaxEnt-RL objective with action chunk size $N$ by introducing the objective 
\begin{equation}
\label{eq::MAxEntChunk}
J(\pi) = \mathbb E_{s_0 \sim \rho_0} V^\pi(s_0),
\end{equation}
where $V^\pi(s_t)$ is defined as
\begin{equation}
\label{eq::Vpi}
V^\pi(s_t) :=\sum_{k=0}^\infty \mathbb{E}_{\rho^\pi}\gamma^{k}\big[r_{t+k} -\alpha \log \pi(a_{t+k}\mid s_{t+N\lfloor k /N \rfloor})\big] .
\end{equation} 
For a given state $s_{t}$, the whole action chunk $a_{t:t+N}$ is sampled from the policy $\pi(a_{t:t+N}|s_{t})$. 
Thus, starting from time step $t$, action $a_{t+k}$ only depends on the last step divisible by $N$, ie. $s_{t+N\lfloor k / N \rfloor}$. The sampled action chunk $a_{t:t+N}$ is then executed in an open-loop fashion in the environment. All intermediate states and values are saved in the replay buffer and used for the critic and actor updates (see Figure~\ref{fig:critic}).
The objective in Equation \ref{eq::MAxEntChunk} generalizes the known MaxEnt-RL objective in Equation~\ref{eq:max_ent_obj} to the action chunking case. 
The entropy bonus is calculated and discounted for every action separately. Therefore, the optimal policy is independent of the chunk size, which would not be the case if the entropy bonus was calculated using the joint likelihood of a chunk's actions. 
The generalization can be easily verified by setting the chunk size to $N=1$, which recovers Objective~\ref{eq:max_ent_obj}.

\textbf{Action Chunk Extended Q-function.} Objective~\ref{eq::MAxEntChunk} naturally leads to an action chunk extended Q-function 
\begin{equation}
\label{eq::MaxEntChunk_Q}
Q^\pi(s_t,a_{t:t+N}) = r_t + \mathbb{E}_{p} \left[\sum_{n=1}^{N-1}\gamma^n r_{t+n} + \gamma^N V^\pi(s_{t+N})\right]
\end{equation}
As opposed to the common definition of the Q-function, here, the action chunk $a_{t:t+N}$ is passed as an argument to the Q-function instead of a single action $a_t$~\cite{li2025qchunk,dong2025chunkingthecritic}. 
These actions are fixed and executed in the environment in an open-loop manner, leading to the state sequence $s_{t+1:t+N+1}$ and the corresponding reward values $r_{t:t+N}$.
We indicate this open-loop execution of the action chunk using the first expectation under the state transition probability $p$.
Please note that this distribution is opposed to the expectation over $\rho^\pi$ in $V^\pi(s_t)$ (Equation~\ref{eq::Vpi}) that explicitly follows from executing the stochastic policy $\pi$ rather than a fixed action sequence.

Intuitively, the extended Q-function measures the expected return when executing the action chunk $a_{t:t+N}$ in state $s_t$ and following the action chunking policy $\pi$ thereafter.
This aligns with the general definition of the Q-function, but extends it to the action chunking case, which can be easily verified by setting the action chunk length $N=1$ and thereby recovering the regular Q-function of MaxEnt-RL (cf. Equation~\ref{eq::MaxEnt_Q}). 

\subsection{Multi-Horizon Q-Function Targets}\label{sec:multi_horizon}
A significant advantage of the Q-function definition in Equation~\ref{eq::MaxEntChunk_Q} is that the temporal structure of the action sequence can be leveraged for off-policy updates without requiring any importance weights \cite{munos2016retracelambda} to account for the bias induced by using off-policy data in the Q-value targets.

More concretely, we define the Q-value target $G^{(N)}$ as 
\begin{align}
\label{eq:n_step_return}
G^{(N)}(s_t,a_{t:t+N}) = \underbrace{\sum_{i=0}^{N-1}\gamma^i\, r_{t+i}}_{\text{N-step return}} + \gamma^N \hat{Q}^{(N)}_{\bar{\phi}}(s_{\tilde t}, a_{\tilde t:\tilde t+N}),
\end{align}
where $\tilde{t}=t+N $ and the target value are computed using target parameters $\bar\phi$ and sampled next-state action chunk $a_{\tilde t:\tilde t+N}\sim \pi_\theta\bigl(\cdot\mid s_{\tilde t}\bigr)$ as
\begin{equation}
    \hat{Q}^{(N)}_{\bar{\phi}}(s_{\tilde t}, a_{\tilde t:\tilde t+N}) = 
    \min_{j\in\{1,2\}} \left[Q_{\bar{\phi}_j}^{(N)}(s_{\tilde t}, a_{\tilde t:\tilde t+N}) - \alpha \sum_{i=0}^{N-1}\gamma^i\log \pi_\theta\bigl(a_{\tilde t+i}\mid s_{\tilde t}\bigr)\right].
\end{equation}
However, simply using $G^{(N)}$ to train the Q-function does not exploit the available data effectively because information is lost
while adding up the chunk's $N$ rewards in Equation~\ref{eq::MaxEntChunk_Q}. Therefore, we train the critic on all target values $G^{(1)}, 
G^{(2)},...,G^{(N)}$ (calculated from the chunk's prefixes) in parallel when updating the network parameters $\phi$. 
As seen in Figure~\ref{fig:critic}, each target value is calculated using Equation~\ref{eq:n_step_return}, but with varying horizons $N$, such that \textit{multi-horizon targets} are calculated.
These multi-horizon targets were also recently proposed by \citet{li2024top}, but in the episode-based RL framework with finite time horizon tasks.
Causal models such as transformers suit this target well, as the $k$-step target $G^{(k)}$ only depends
on the first $k$ actions of a chunk.
We therefore follow \citet{li2024top} and use a causal transformer critic to effectively handle multi-horizon targets.
The final objective for updating the parameters $\phi$ is then given by 
\begin{equation}
\label{eq:Jphi_mse}
    J(\phi) = \frac{1}{2N}\sum_{k=1}^{N}\Big( Q^{(k)}_\phi(s_t, a_{t:t+k}) - G^{(k)}(s_t, a_{t:t+k}) \Big)^2,
\end{equation}
where the data is sampled from the replay buffer $\mathcal{B}$.
Figure~\ref{fig:critic} visualizes the critic update.

\subsection{Optimizing the Action Chunking Policy}
With the Q-function $Q_\phi$ from Section \ref{sec:multi_horizon} we can easily update the chunking policy $\pi_\theta(a_{t:t+N}|s_t)$ following recent MaxEnt-RL approaches \cite{haarnoja2017reinforcement, haarnoja2018soft} by using the reverse KL divergence 
\begin{equation}
\label{eq:Jphi}
    \begin{split}
    J(\pi) &= \mathbb{E}_{\mathcal{B}}\Big[
    \operatorname{KL}\Big(
    \pi_\theta(\cdot \mid s_t)\;\Big\|\;
    \frac{\exp\!\big(Q^{(N)}_\phi(s_t,a_{t:t+N})/\alpha\big)}{Z(s_t)} \Big) \Big],
    \end{split}
\end{equation}
where the states $s_t$ are sampled from the replay buffer $\mathcal{B}$.

\subsection{State Coverage in Action Chunking RL}
\begin{figure}
    \centering
    \hfill
    \begin{subfigure}[t]{.21\textwidth}
        \centering
        \vbox to 3.7cm {\vfill \includegraphics[width=\linewidth]{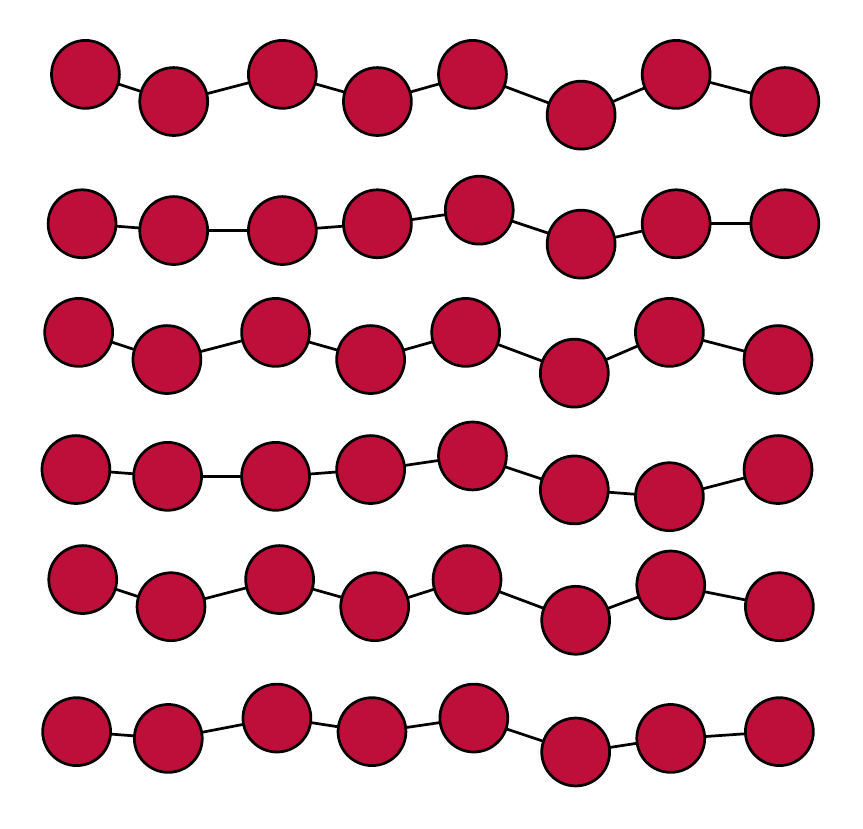} \vfill}
        \caption{\centering \\No Action Chunking}
        \hfill
        \label{fig:stepbased}
    \end{subfigure}
    \hfill
    \begin{subfigure}[t]{.21\textwidth}
        \centering
         \vbox to 3.7cm {\vfill \includegraphics[width=\linewidth]{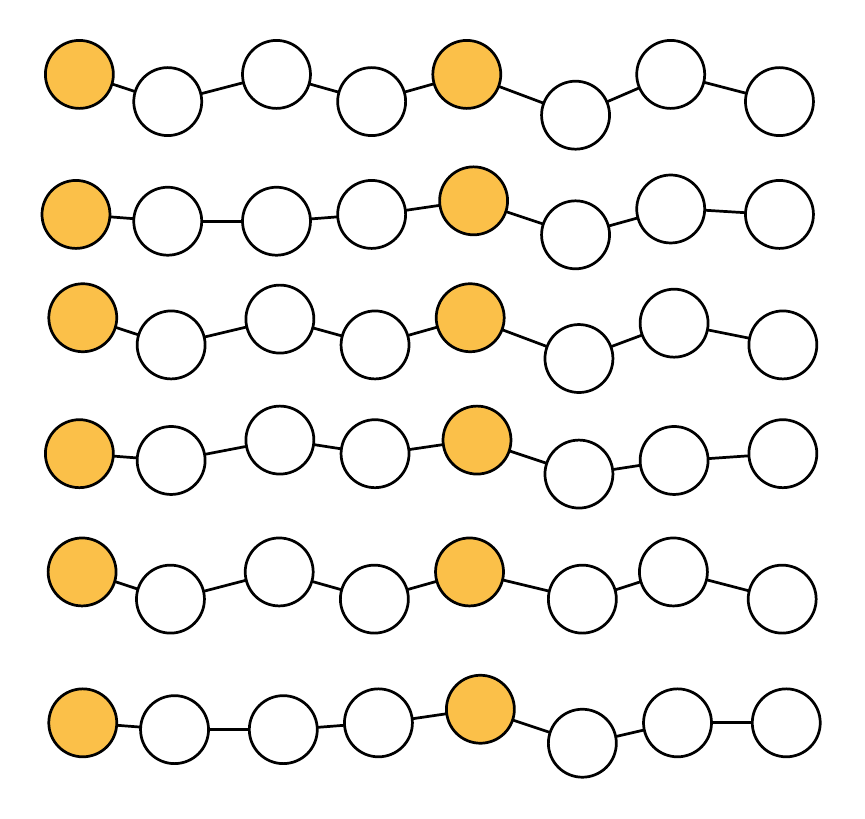} \vfill}
        \caption{\centering Uniform \\ Action Chunking}
        \label{fig:replan_naive}
    \end{subfigure}
    \hfill
    \begin{subfigure}[t]{.21\textwidth}
        \centering
         \vbox to 3.7cm {\vfill \includegraphics[width=\linewidth]{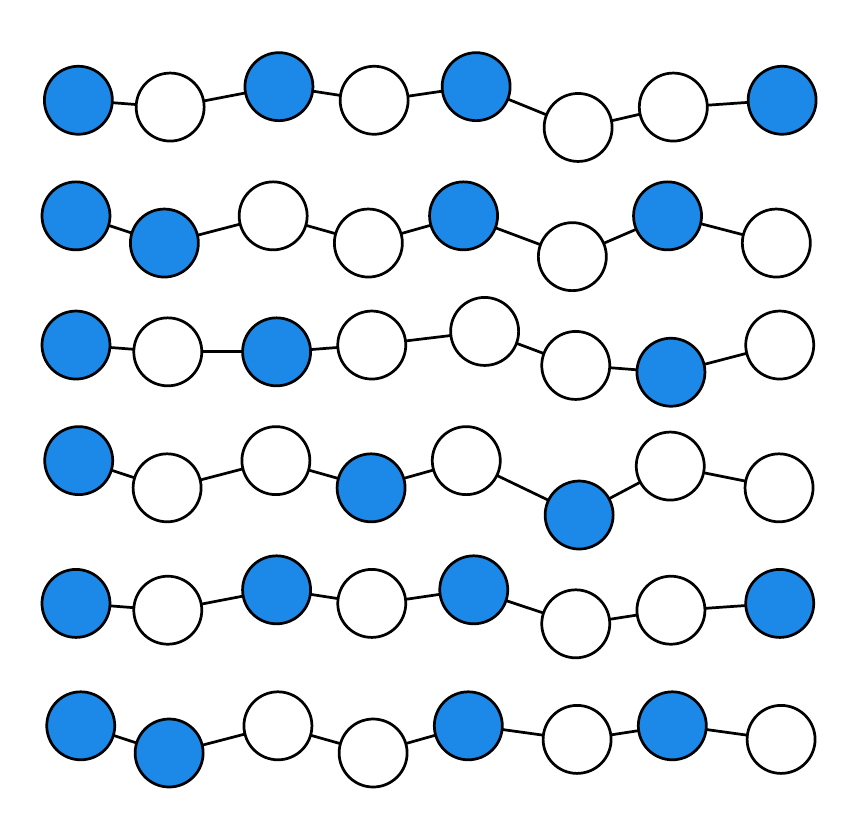} \vfill}
        \caption{\centering SEAR's \\ Random Replanning}
        \label{fig:replan_random}
    \end{subfigure}
    \hfill
    \begin{subfigure}[t]{.32\textwidth}
        \centering
        \includegraphics[width=\linewidth]{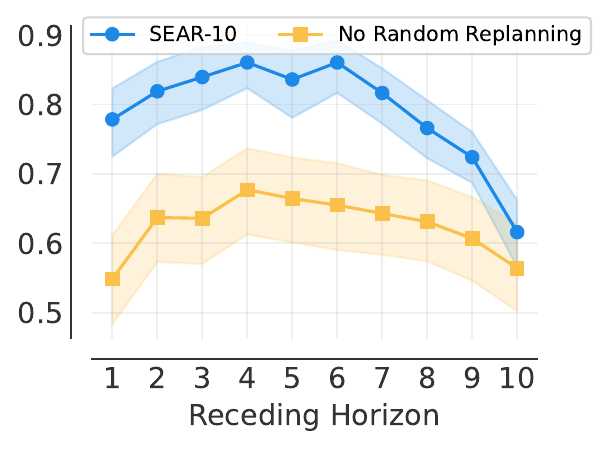}
        \caption{Metaworld-Hard\\ IQM Success}
        \label{fig:replan_result}
    \end{subfigure}
    \hfill
    \caption{
        \textbf{Uniform Action Chunking Limits Actor State Coverage.} In methods without
action chunking like T-SAC and SimbaV2, the actor is evaluated in each
state of the trajectory \textbf{(a)}. Prior action chunking methods evaluate the actor in every N-th state \textbf{(b)}. This creates areas of
the state space where the actor is rarely evaluated during training (white dots). When
evaluating such a policy with receding horizons, the actor is queried
in these out-of-distribution states and the resulting performance drop
offsets most gains from the higher decision frequency. SEAR introduces
\textit{random replanning} which breaks this pattern, restoring uniform
state coverage \textbf{(c)}. As a result, SEAR benefits from receding
horizon control to a much higher degree than methods without random replanning \textbf{(d)}. 
}
    \label{fig:random_init}
\end{figure}

Like most action chunking RL methods, we split the data of the replay buffer into new chunks for the policy and
actor update. The policy and critic are thus updated on chunks that may start in arbitrary time steps. 
However, an action chunking policy with fixed chunk size $N$ is only evaluated at every $N$-th state of the underlying MDP.
Particularly in environments with deterministic or narrowly distributed initial states and large chunk sizes, it can happen that the policy is rarely evaluated on large parts of the state space.
This effect is illustrated in Figure~\ref{fig:replan_naive} for $N=4$, where colored circles denote replanning states and white circles indicate intermediate states visited only during open-loop execution.
As shown in Section~\ref{sec:results}, we identify this as a key limitation of fixed-size action chunking. To address it, we propose \textit{random replanning}: During data collection, we only execute a random prefix of each chunk. This increases the actor's state coverage, leading to more reliable updates for both the critic and the actor.

\section{Experiments}
\label{sec:results}

We evaluate SEAR in two settings: purely online RL on Metaworld and offline-to-online RL
on OGBench. We compare against state-of-the-art baselines in both settings and conduct
ablation studies to understand the contribution of each design choice.

\begin{figure*}[t]
\centering
    \hfill
    \begin{subfigure}[b]{.49\textwidth}
        \centering
        \includegraphics[width=\linewidth]{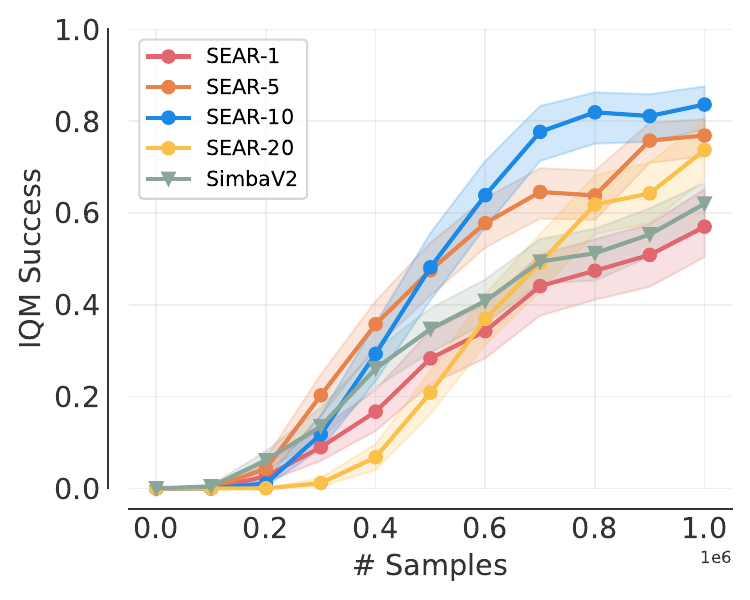}
        \caption{Sample Efficiency}
        \label{fig:ablations-chunksize-efficiency}
    \end{subfigure}
    \hfill
    \begin{subfigure}[b]{.49\textwidth}
        \centering
        \includegraphics[width=\linewidth]{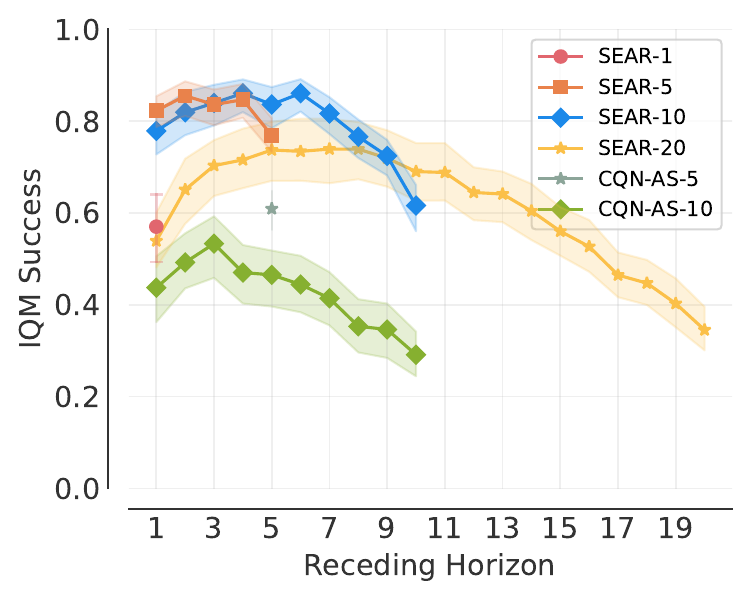}
        \caption{Effect of receding horizons }
        \label{fig:ablations-chunksize-replan}
    \end{subfigure}
    \hfill
\caption{
    \textbf{SEAR works well across Chunk Sizes and benefits consistently from Receding Horizons.} We compare IQM Success on Metaworld-Hard for different chunk sizes. 
\textbf{(a)} All SEAR policies with action chunking achieve higher performance than SimbaV2, with only the single-action version SEAR-1 having marginally lower performance.
The best performance is achieved by SEAR-10, indicating a favourable trade-off between value propagation and decision frequency.
\textbf{(b)} For all tested chunk sizes, including one (i.e., no action chunking), training a SEAR policy on the next larger chunk size and using it with a receding horizon
yields higher performance than using a policy with a smaller chunk size and no receding horizon. All tested policies degrade in performance when increasing
the receding horizon towards the chunk size, highlighting the challenge of reduced decision frequencies. Similarly, all action chunking policies drop in performance
    when reducing the receding horizon towards one. The optimal receding horizon grows with the chunk size, from two for SEAR-5 to seven to eight for SEAR-20.
}
\label{fig:ablations-chunksize}
\end{figure*}

\subsection{Experiment Setup}

We evaluate SEAR in the online setting on the Metaworld
MT1 Benchmark~\cite{yu2021metaworldbenchmarkevaluationmultitask,mclean2026metaworld},
which features a diverse set of 50 manipulation tasks. We
follow~\citet{agarwal2021deep} and report the IQM success
and bootstrap 95\% confidence intervals. Following prior work on
Metaworld~\cite{dong2025chunkingthecritic,li2024top,li2024latent,joshi2025benchmarkmp},
we only consider episodes to be successful if the success condition holds
in the episode's last time step. This increases the task complexity
and forces longer-term and stable behavior. Even with these stricter
success criteria, many Metaworld tasks remain trivially solvable. We therefore focus on the
20 hardest tasks in Metaworld (Metaworld-Hard), selected based on SimbaV2's performance in preliminary experiments.
For transparency, the IQM results for all 50 tasks are included in
Appendix~\ref{app:extended_results}. On Metaworld, all SEAR experiments were conducted with 10 seeds, while SimbaV2 and CQN-AS were evaluated on
5 seeds.

In the offline-to-online setting, we use the first four tasks of the
OGBench~\cite{ogbench_park2025} cube-triple-play domain, as they are the hardest
manipulation tasks that QC and QC-FQL can solve with a dataset of 1M offline
transitions~\cite{li2025qchunk}. All OGBench results were conducted with 8 seeds and
include 1 million offline steps before transitioning to the online phase. We apply
SEAR's multi-horizon critic targets and random replanning to the QC framework while
keeping all other hyperparameters and network architectures fixed. Because QC queries
the critic extensively, especially when combined with multi-horizon targets, we use an MLP
critic rather than a transformer to keep wall-clock training times feasible.

We test SEAR with chunk sizes of 5, 10, and 20, and include chunk size 1 as a
single-action baseline. As online RL baselines we use CQN-AS~\cite{seo2024coarse} with
chunk sizes 5 and 10 and SimbaV2~\cite{lee2025simbav2}, representing the state of the
art with and without action chunking respectively. Further baselines (T-SAC, SAC,
TOP-ERL and a naively chunked SimbaV2) are described in
Appendix~\ref{app:exp_details:fig1}. Unless otherwise noted, we report
success after 1 million online environment steps.

All action chunking policies were evaluated using a receding
horizon length of 5. For chunk size 5 policies, this has no
effect, easing comparison to prior methods without receding
horizons. Implementation details and hyperparameters are provided in
Appendix~\ref{app:exp_details}.

\begin{figure}[t]
    \centering
    \begin{subfigure}[b]{.49\textwidth}
        \centering
        \includegraphics[width=\linewidth,keepaspectratio]{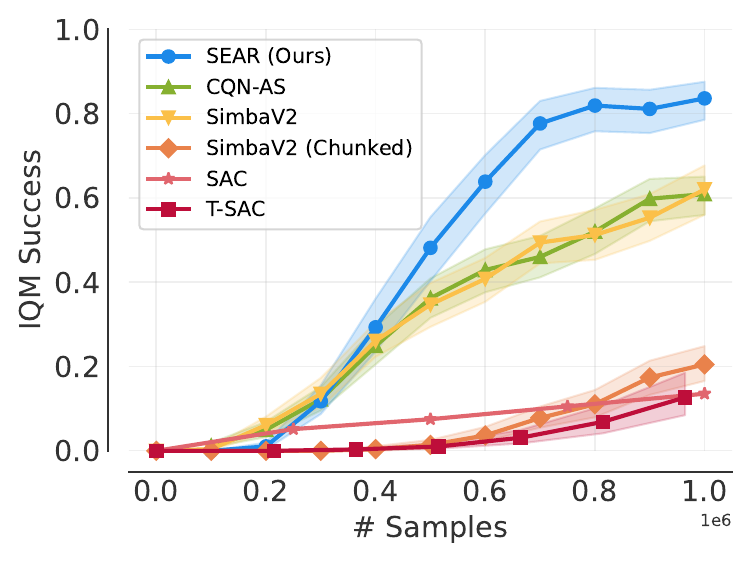}
        \caption{Metaworld-Hard}
        \label{fig:metaworld_comparison}
    \end{subfigure}
    \hfill
    \begin{subfigure}[b]{.49\textwidth}
        \centering
        \includegraphics[width=.98\linewidth]{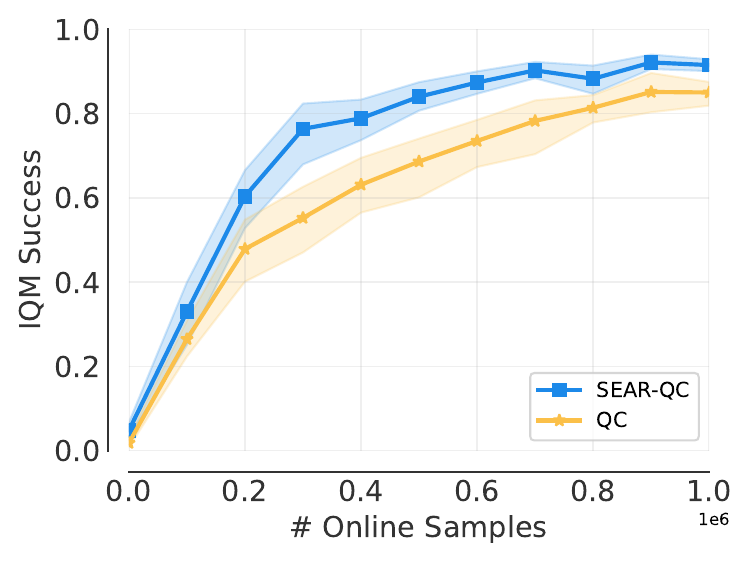}
        \caption{OGBench Cube-Triple-Play 1M}
        \label{fig:ogbench}
    \end{subfigure}
\caption{
    \textbf{Comparison to State-of-the-Art Baselines.} \textbf{(a)} In the online setting, SEAR achieves over 80\% IQM success, significantly above the 60\% success of CQN-AS and SimbaV2, which we consider to be the state-of-the-art in online RL with and without action chunking respectively. As further baselines we include T-SAC which shares a similar critic with SEAR, SAC and a naive action chunking extension of SimbaV2. They achieve at most 21\% IQM success. \textbf{(b)} In the offline-to-online setting, we achieve higher IQM success rates than QC, especially early in training.
}
    \label{fig:comparison}
\end{figure}

\subsection{Comparison to State-of-the-Art}
\label{sec:comparison}

Figure~\ref{fig:comparison}a shows that SEAR achieves over 80\% IQM success on Metaworld-Hard,
substantially outperforming SimbaV2 and CQN-AS, the state of the art in online RL without
and with action chunking respectively, which both reach around 60\%.
A naive chunked extension of SimbaV2 achieves only around 20\% IQM success, demonstrating
that the action chunking techniques common in offline and offline-to-online RL do not
transfer to the online setting without further algorithmic work.
T-SAC, which chunks the critic but retains a single-step actor, performs similarly poorly.

In the offline-to-online setting (Figure~\ref{fig:comparison}b), applying SEAR's components
to the QC framework outperforms QC across the OGBench cube-triple-play
tasks, with the advantage most pronounced early in online training.

\subsection{Why Does SEAR Outperform Prior Action Chunking Methods?}
\label{sec:ablations}

We attribute SEAR's gains to two design choices that address distinct failure modes of
prior action chunking methods.

\paragraph{Actor state coverage.}
Figure~\ref{fig:replan_result} demonstrates that random replanning is necessary for effective use of
receding horizon control. Without it, SEAR benefits less from shorter receding
horizons (from 56\% to 68\% IQM success, compared to 62\% to 86\% with random replanning). Figure~\ref{fig:ablations-chunksize-replan} shows the same failure mode in
prior methods: CQN-AS-10 with a receding horizon does not match CQN-AS-5 without one.

\paragraph{Critic design.}
The ablation in Figure~\ref{fig:critic} shows that multi-horizon targets are the
most important component of SEAR's critic, as removing them and fitting only the full-chunk
target $Q^{(N)}$ reduces performance to around 36\% IQM success. Bootstrapping the transformer critic with action chunks rather than
single-step actions also matters, as it allows the multi-horizon targets to improve value
propagation. Replacing the transformer backbone with an MLP reduces performance, but
SEAR with an MLP remains competitive, confirming that the transformer alone does not
account for the gains.

\paragraph{Receding horizon analysis.}
Figure~\ref{fig:ablations-chunksize-efficiency} shows that all SEAR policies with action
chunking outperform SimbaV2, with SEAR-10 achieving the best final performance. SEAR-1
(no action chunking) performs slightly below SimbaV2, indicating that the use of a transformer critic is not 
the source of SEAR's performance gains.
Figure~\ref{fig:ablations-chunksize-replan} shows that a medium-size receding horizon is optimal.
Horizons near the full chunk size reduce decision frequency and harm
reactivity.
For horizons near one, we frequently observed pausing behavior of the policy, leading to low success rates. SEAR
policies achieve maximum performance at intermediate receding horizons: two for SEAR-5, four to six for SEAR-10 and seven to eight for SEAR-20.
Notably, SEAR-5 and SEAR-10 deployed at a receding
horizon of one outperform both SEAR-1 and SimbaV2, making action chunking with SEAR
an effective way to obtain a high-performing single-action policy.

\section{Conclusion}
\label{sec:conclusion}

This work introduced SEAR, an action chunking algorithm for the previously
underexplored online RL setting, achieving over 80\% IQM success rate on
Metaworld-Hard and substantially outperforming both action chunking and
non-action chunking baselines.

The core of SEAR's design is that multi-horizon critic targets provide a proper
critic fit over the enlarged action space of chunks. 
Given a well-fitted critic, random replanning restores the state
coverage that open-loop chunk execution degrades, making receding horizon control
effective. Together, these two components allow SEAR to benefit from the fast
value propagation and structured exploration of long chunks during training,
while maintaining the reactivity of a short decision horizon at evaluation time.
Through extensive ablation studies, we demonstrated that both components are necessary.

We further demonstrated that SEAR's multi-horizon targets and random replanning
transfer to the offline-to-online setting. Applying them to the QC framework,
while keeping all other components fixed, yields consistent improvements on the
OGBench cube-triple-play tasks. This suggests that the core ideas behind SEAR are not specific
to the online setting and can benefit action chunking methods more broadly.

\paragraph{Limitations}
Action chunking RL is best suited to sufficiently long-horizon tasks that do not require high reactivity, such as robotic manipulation, where action chunking aids exploration and task horizons are long enough to benefit from improved value propagation.
Current action chunking methods do not work in high-reactivity domains like locomotion, as action chunking, even when only used during training, degrades data quality.

As such, there are few benchmarks available for which action chunking is a good fit.
In this work, we focus on Metaworld~\cite{yu2021metaworldbenchmarkevaluationmultitask} for online RL, which offers the currently broadest set of diverse manipulation tasks for state-based online RL.
Other benchmarks such as ManiSkill~\cite{gu2023maniskill2} operate at very short episode horizons.
In the offline-to-online setting, widely used benchmarks such as Robomimic~\cite{robomimic2021} and most OGBench domains~\cite{ogbench_park2025} are saturated, limiting insightful evaluation to a narrow set of tasks. 
Furthermore, our experiments are limited to simulation. As a step towards a more realistic setting, we provide vision results in Appendix~\ref{vision}.

SEAR incurs a higher computation cost during training than single-step methods due to the multi-horizon targets and the transformer critic.
Per environment step, SEAR-10's network updates take roughly twice as long as SimbaV2's, so under a matched update budget, SimbaV2 at 1M steps (62\% IQM success) outperforms SEAR-10 at 500k steps (48\%). SEAR is therefore most attractive when environment interaction rather than compute is the bottleneck, as is common in robotics.
However, compared to the state-of-the-art action chunking RL method CQN-AS~\cite{seo2024coarse}, policy evaluation is fast, as action chunks are generated with a single neural network forward pass. We analyze this in further detail in Appendix~\ref{computational-efficiency}.

\paragraph{Outlook}
To make action chunking methods more broadly useful, including in domains requiring high reactivity such as locomotion, dynamic replanning algorithms would enable training and evaluation with an action chunk size that adapts to the current observation.
Random replanning and receding horizon control should also transfer to on-policy RL, where an MLP critic with multi-horizon targets may offer a favorable trade-off between sample efficiency and wall-clock time. Recent actor-critic approaches for on-policy RL~\cite{voelcker2026relative} are particularly promising candidates for this.

In addition, the computational overhead of the transformer critic could be reduced with more efficient sequence models such as state space models~\cite{lahoti2026mamba} that keep the causal structure at a computational cost that is linear in the number of tokens.

\begin{ack}
The research presented in this paper was funded by the Horizon
Europe Project XSCAVE under Grant 101189836. This work was further
supported by the European Research Council (ERC) under the European
Union’s Horizon Europe programme through the project SMARTI$^3$ (Grant
Agreement No. 101171393). The authors acknowledge support by the state
of Baden-Württemberg through bwHPC, and the HoreKa supercomputer. This
work has been supported by the German Federal Ministry of Research,
Technology, and Space (BMFTR) under the Robotics Institute Germany (RIG).
The authors gratefully acknowledge the Gauss Centre for Supercomputing e.V. (www.gauss-centre.eu) for funding this project by providing computing time on the GCS
Supercomputer JUWELS at Jülich Supercomputing Centre (JSC).
\end{ack}

\clearpage

\bibliography{iclr2026_conference}
\bibliographystyle{icml2026}

\clearpage

\appendix

\section{Algorithm}
\label{app:algorithm}
Algorithm~\ref{alg:sear} summarizes SEAR's training procedure.
\begin{algorithm}[H]
\caption{Sample Efficient Action Chunking RL (SEAR)}
\label{alg:sear}
\begin{algorithmic}[1]
\STATE Initialize actor parameters $\theta$, critic parameters $\phi_1, \phi_2$, target critic parameters $\bar{\phi}_1 \leftarrow \phi_1, \bar{\phi}_2 \leftarrow \phi_2$, empty replay buffer $\mathcal{B}$
\FOR{step $= 1, 2, \ldots$}
    \STATE \textcolor{gray}{\# Data collection with random prefix execution}
    \IF{$|\mathcal{B}| <$ step}
        \STATE Sample action chunk $a_{t:t+N} \sim \pi_\theta(\cdot \mid s_t)$
        \STATE Sample random prefix length $k \sim \text{Uniform}\{1, \dots, N\}$
        \FOR{$i = 0$ to $k-1$}
            \STATE Execute $a_{t+i}$ in environment, observe $r_{t+i}, s_{t+i+1}$
            \STATE Add transition $(s_{t+i}, a_{t+i}, r_{t+i}, s_{t+i+1})$ to $\mathcal{B}$
        \ENDFOR
    \ENDIF

    \STATE \textcolor{gray}{\# Update Networks}
    \STATE Sample batch of chunks $(s_t, a_{t:t+N}, r_{t:t+N}, s_{t+N})$ from $\mathcal{B}$
    \STATE Update critics by minimizing $J(\phi_{1,2})$ \hfill (Equation~\ref{eq:Jphi_mse})
    \STATE Update actor by minimizing $J(\theta)$ \hfill (Equation~\ref{eq:Jphi})

    \STATE update $\alpha$ with $\mathcal H_\text{target} = -N \cdot \dim(\mathcal A)$ \hfill \cite{haarnoja2018soft}
    \STATE \textcolor{gray}{\# Update target networks}
    \STATE $\bar{\phi}_j \leftarrow \tau \phi_j + (1-\tau) \bar{\phi}_j$ for $j \in \{1,2\}$
\ENDFOR
\end{algorithmic}
\end{algorithm}

\section{Vision Experiments}
\label{vision}
To test how SEAR handles vision observations, we trained SEAR-10 and
SEAR-1 on six Metaworld tasks with visual observations: hammer-v3,
handle-pull-v3, handle-pull-side-v3, peg-unplug-side-v3,
plate-slide-side-v3 and door-open-v3.

We chose these tasks because both SEAR-10 and SEAR-1 solved them within 500k steps
and SimbaV2 reached at least 50\% IQM success on each of them within 500k steps in the state-based
setting (Appendix Figure~\ref{fig:grid}), which let us obtain a meaningful,
non-floor/non-ceiling comparison. To isolate the effects of SEAR's action 
chunking, we took care not to choose tasks where SEAR-10 already compares 
favorably to SEAR-1 with state observations. Following recent work on 
real-world, vision-based RL~\cite{luo2025hilserl}, we encoded the observations 
using the pretrained ResNet-50 encoder presented in~\cite{nair2022r3m}.
The encoder is kept frozen during training. At every step, we render a single
RGB frame from Metaworld's \texttt{corner} camera, resize it to $256$ pixels,
center-crop it to $224\times224$ and encode it into a $2048$-dimensional
embedding. The policy and critic observe this embedding concatenated with the
4-dimensional proprioceptive state (end-effector position and gripper
opening); we do not use frame stacking. All methods use the same observations
and the hyperparameters from Table~\ref{tab:hyperparams}.

\begin{figure}[t]
    \centering
    \begin{subfigure}[b]{.49\textwidth}
        \centering
        \includegraphics[width=\linewidth,keepaspectratio]{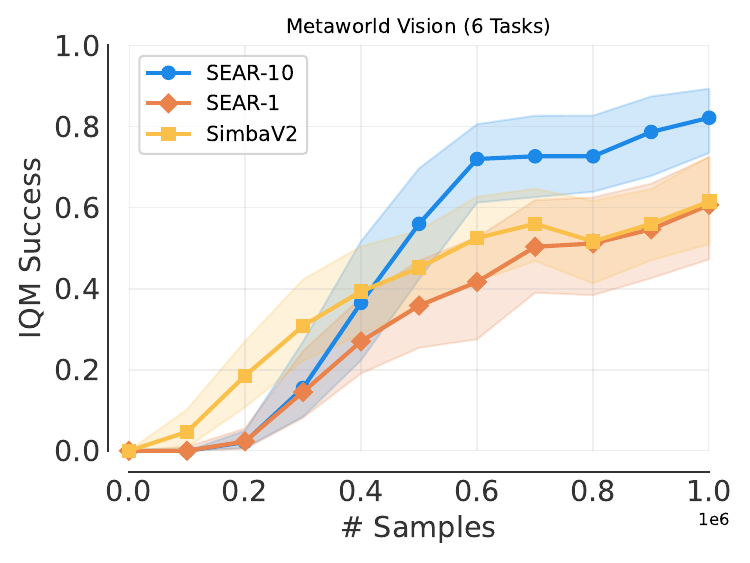}
        \caption{Metaworld-Vision}
        \label{fig:vision_comparison}
    \end{subfigure}
    \hfill
    \begin{subfigure}[b]{.49\textwidth}
        \centering
        \includegraphics[width=.98\linewidth]{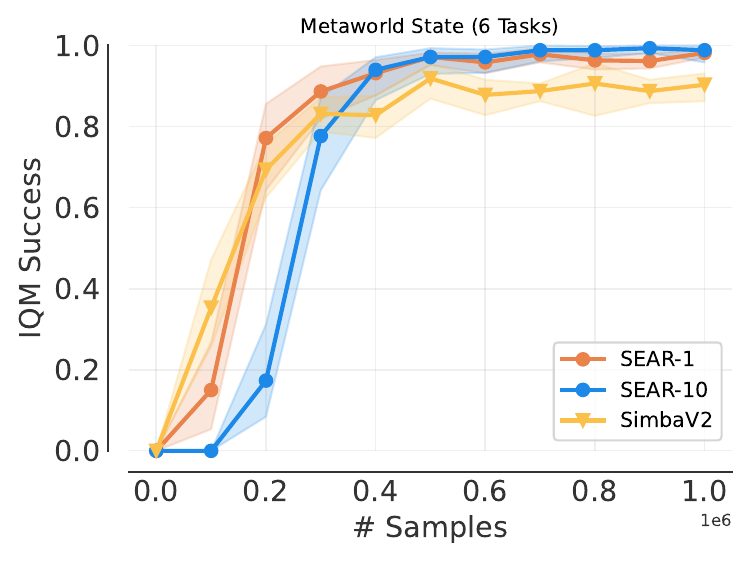}
        \caption{State-based Results on the same Tasks.}
        \label{fig:vision_state}
    \end{subfigure}

\caption{
    \textbf{SEAR-10 works well with visual observations.} \textbf{(a)} Across six visual manipulation tasks, SEAR-10 outperforms both its non-action chunking variant SEAR-1, as well as the non-chunking baseline SimbaV2. 
    \textbf{(b)} With state observations, SEAR-10 does not outperform SEAR-1 on the tasks chosen for visual manipulation. This makes it unlikely that the vision results are explained by the specific environments tested. The per-task training curves are provided in
Figure~\ref{fig:visual_grid}.
}
    \label{fig:vision}
\end{figure}

As seen in Figure~\ref{fig:vision}, SEAR-10 has a distinct advantage over SEAR-1
in both sample efficiency and final performance when using visual observations. At 1M online steps, SEAR-10 has an IQM success of 82\% vs SEAR-1 with 61\% (10 seeds per task).
Compared to both SEAR variants, SimbaV2 has the advantage until around 400k environment steps, but the final performance at 1M steps is comparable to SEAR-1 at 62\% IQM success. This suggests that SEAR's action chunking is effective at visual manipulation. The per-task training curves are provided in
Figure~\ref{fig:visual_grid}.

\section{Implementation Details}
\label{app:implementation}

Code, training curves and interactive visualizations can be found at \url{https://alrhub.github.io/SEAR/}.

\section{Experiment Details}
\label{app:exp_details}

\subsection{Compute Demands}
\label{app:compute}

All experiments were performed on NVIDIA A100 GPUs.

\paragraph{OGBench.}
We evaluate three methods across 4 tasks with 8 seeds each:
QC-SEAR ($11\,\text{h} \times 8 \times 4 = 352$ GPU hours),
QC ($4\,\text{h} \times 8 \times 4 = 128$ GPU hours), and
QC-FQL ($3\,\text{h} \times 8 \times 4 = 96$ GPU hours).
Subtotal: 576 GPU hours.

\paragraph{Metaworld.}
SEAR main results (4 chunk sizes, 50 tasks, 10 seeds):
$3\,\text{h} \times 10 \times 4 \times 50 = 6{,}000$ GPU hours.
SEAR ablations (3 ablations, 20 tasks, 10 seeds):
$3\,\text{h} \times 10 \times 3 \times 20 = 1{,}800$ GPU hours.
CQN-AS (2 chunk sizes, 20 tasks, 5 seeds):
$16\,\text{h} \times 5 \times 2 \times 20 = 3{,}200$ GPU hours.
SimbaV2 (50 tasks, 5 seeds):
$2\,\text{h} \times 5 \times 50 = 500$ GPU hours.
Subtotal: 11,500 GPU hours.

\textbf{Total: 12,076 GPU hours.}

For SEAR, four seeds are run in parallel on a single GPU, so the 3h per seed above reflects $12\,\text{h} / 4$ wall-clock time rather than the runtime of a single training run. We run multiple seeds in parallel as the extensive evaluation,
as well as our current replay buffer implementation limit GPU utilization.
These numbers represent a conservative estimate of the required compute, as we used the runtime of SEAR-10 for all SEAR variants, while SEAR-1 and SEAR-5 train faster. Additionally, the run times of SEAR and CQN-AS include evaluation across all possible receding horizons, which takes significant time. The above does not include compute spent developing the method and debugging its implementation.

\subsection{Assets and Licenses}
\label{app:licenses}
We use the following existing assets:
Metaworld~\cite{yu2021metaworldbenchmarkevaluationmultitask,mclean2026metaworld} (MIT License),
OGBench~\cite{ogbench_park2025} (MIT License),
the MuJoCo physics engine~\cite{todorov2012mujoco} underlying both benchmarks (Apache License 2.0),
the pretrained R3M encoder~\cite{nair2022r3m} (MIT License), and
the QC implementation~\cite{li2025qchunk} (MIT License).
From SimbaV2~\cite{lee2025simbav2} (Apache License 2.0), we use the MLP implementation
within our training code for the main experiments and the reference implementation
for the comparison in Figure~\ref{fig:simbav2_diff_implementations}.
The CQN-AS implementation~\cite{seo2024coarse} is publicly available without an
explicit license; we use it for research purposes only.

\subsection{SEAR}
\label{app:sear_hparams}

As for all action chunking methods, the chunk size is a key hyperparameter of SEAR.
Our results suggest a chunk size of 10 as a default, with 5 as an alternative if
training with 10 is unsuccessful (cf.~Figure~\ref{fig:ablations-chunksize}a).
On Metaworld, whose base control step is 12.5\,ms (80\,Hz), a chunk of 10 actions
corresponds to only 125\,ms, whereas imitation learning typically uses chunks
spanning roughly 1--2\,s~\cite{zhao2023act,chi2024diffusionpolicy}.
Since SEAR randomizes the executed prefix during data collection, the receding
horizon is not fixed at training time and can be tuned at evaluation time without
retraining (cf.~Figures~\ref{fig:ablations-chunksize}b and~\ref{fig:replan_result}).
The best receding horizon grows with the chunk size: two steps for SEAR-5, four to six for SEAR-10 and seven to eight for SEAR-20 on Metaworld-Hard.

SEAR's main hyperparameter besides the chunk size is the embedding dimension of the transformer (``hidden dimension'' in Table~\ref{tab:hyperparams}). We tried 128, 256 and 512 and chose 512 as the smaller ones had insufficient capacity. With regard to loss functions we tried $\ell_2$ and C51 and chose the latter because it stabilized training. All other hyperparameters were chosen with common values from prior methods and not specifically optimized.

\subsection{SimbaV2}

\begin{figure}[ht]
    \centering
    \includegraphics[width=0.5\linewidth]{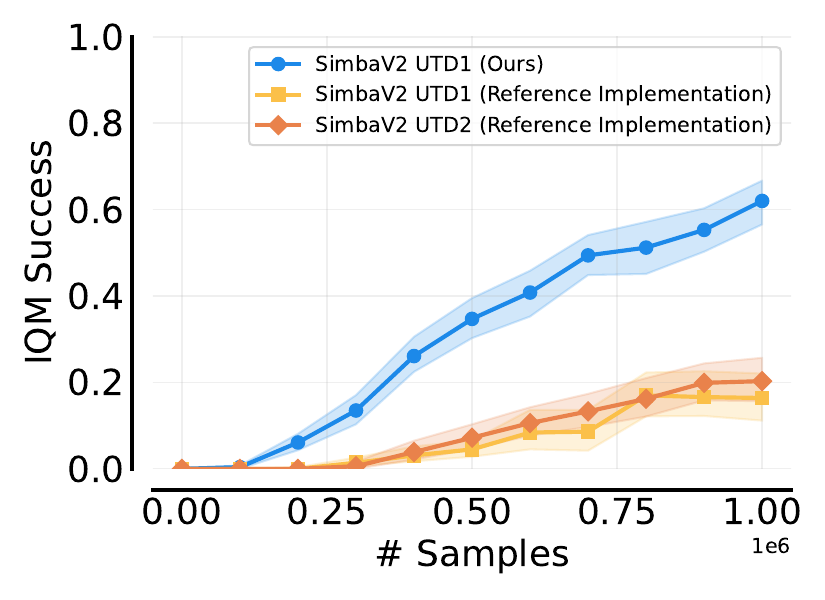}
    \caption{Comparison between our SimbaV2 implementation and the reference implementation.}
    \label{fig:simbav2_diff_implementations}
\end{figure}

For SimbaV2, we use our reimplementation within the same training pipeline as SEAR and disable the reference reward scaling, since Metaworld rewards are already on a consistent scale across tasks, while keeping all other components and techniques described in Section 4 of SimbaV2~\cite{lee2025simbav2}. For transparency, we also report results with the original implementation, shown in Fig.~\ref{fig:simbav2_diff_implementations}.

\subsection{Hyperparameter values}
Table~\ref{tab:hyperparams} gives an overview of the hyperparameters that we use for SEAR and baselines.
Below, we provide further details on the specific hyperparameter choices. The full hyperparameters and
algorithm-specific options can be found in our code release.

\begin{table}[h]
    \centering
    \caption{Hyperparameters used in our experiments. CQN-AS has no actor component.
    \textsc{SimbaV2-MLP} denotes the MLP head used in SimbaV2, including its spherical normalization. $\dim(A)$ denotes the dimensionality of a single action, and $N$ is the chunk size.}
    \label{tab:hyperparams}
    \begin{tabular}{llll}
        \toprule
        \textbf{Component} & \textbf{SEAR} & \textbf{CQN-AS} & \textbf{SimbaV2} \\
        \midrule
        \multicolumn{3}{l}{\textit{Actor}} \\
        \quad architecture & SimbaV2-MLP & \text{N/A} & SimbaV2-MLP\\
        \quad hidden dimension & 512 & \text{N/A} & 512 \\
        \quad number of blocks & 1 & \text{N/A} & 2\\
        \midrule
        \multicolumn{3}{l}{\textit{Critic}} \\
        \quad architecture & Transformer & MLP + GRU & SimbaV2-MLP \\
        \quad number of bins & 101 & 101 & 101 \\
        \quad value range & $[0, 1000]$ & $[0, 1000]$ & $[0, 1000]$\\
        \quad hidden dimension & 512 & 1024 & 512\\
        \quad number of heads & 16 & \text{N/A} & \text{N/A}\\
        \quad number of blocks & 2 & 1 & 2\\
        \midrule
        \multicolumn{3}{l}{\textit{Algorithm}} \\
        \quad total timesteps & $1{,}000{,}000$ & $1{,}000{,}000$ & $1{,}000{,}000$\\
        \quad seed timesteps & $0$ & $4{,}000$ & $0$\\
        \quad update-to-data ratio (UTD) & 1 & 1 & 1\\
        \quad chunk size & 1 / 5 / 10 / 20 & 5 / 10 & 1\\
        \quad batch size & 256 & 256 & 256\\
        \quad discount $\gamma$ & $0.99$ & $0.99$ & $0.99$\\
        \quad target critic update ratio & $5\cdot 10^{-2}$ & $1.0$ & $5\cdot 10^{-2}$ \\
        \quad target entropy & $-N \cdot \dim(A)$ & \text{N/A} & $-\dim(A)$ \\
        \midrule
        \multicolumn{3}{l}{\textit{Optimization}} \\
        \quad critic optimizer & AdamW & AdamW & AdamW\\
        \quad learning rate & $3 \cdot 10^{-4}$ & $5 \cdot 10^{-5}$ & $3 \cdot 10^{-4}$\\
        \quad weight decay & $1\cdot 10^{-4}$ & $0.1$ & $1\cdot 10^{-4}$\\
        \quad $\beta_1$ & $0.9$ & $0.9$ & $0.9$\\
        \quad $\beta_2$ & $0.999$ & $0.999$ & $0.999$\\
        \bottomrule
    \end{tabular}
\end{table}

\paragraph{CQN-AS}
As the authors of \cite{seo2024coarse} neither evaluate on nor provide hyperparameters for Metaworld \cite{yu2021metaworldbenchmarkevaluationmultitask}, we tuned the hyperparameters of CQN-AS specifically for the 20 hardest Metaworld tasks.
Note that, since we operate in a purely online RL setting, we do not use the behavior cloning component of CQN-AS.
We use an exploration noise schedule that linearly decreases from $0.1$ to $0.01$ until step $500{,}000$, and is then kept constant for the remaining training time.
We also adopt the dueling Q-networks and distributional critic that \cite{seo2024coarse} use only for vision-based environments, as we found it to improve performance.
To align CQN-AS with SEAR, we fix the update-to-data at one.
We further tuned the learning rate and batch size, although both had only a minor impact on the final success rate in our experiments.
All other hyperparameters were adopted from the state-based tasks of \cite{seo2024coarse}.
We found that a larger critic or using more bins or levels for the coarse-to-fine action selection algorithm does not improve the success rate on Metaworld.
See Table~\ref{tab:hyperparams} for a summary of the hyperparameters.

\subsection{Sample Efficiency Comparison of SEAR, SimbaV2, and CQN-AS}
\label{app:exp_details:fig1}
Here, we provide further details on the experimental setup for the comparison of the sample efficiency of SEAR and baseline methods in Figure~\ref{fig:comparison}.
The naively chunked SimbaV2 baseline uses action chunking for the actor and critic, but without multi-horizon returns and random replanning.
SEAR and the naively chunked SimbaV2 baseline use a chunk size of 10 with a receding horizon of 5 during evaluation, which performs best according to the ablations in Section~\ref{sec:ablations}. 
CQN-AS uses a chunk size of 5, which performed better than 10 in our experiments, and no receding horizon control.

\paragraph{Additional baselines.}
Figure~\ref{fig:comparison}a further includes T-SAC~\cite{dong2025chunkingthecritic},
which uses a causal transformer critic with N-step returns but a single-step actor,
and SAC~\cite{haarnoja2018soft}. The authors of T-SAC kindly provided the T-SAC
results, and the authors of TOP-ERL~\cite{li2024top} kindly provided the SAC and
TOP-ERL results, each over 8 seeds and with the same last-step success criterion.
For TOP-ERL, only the aggregate IQM success over all 50 Metaworld tasks is
available. For SAC, only per-task IQM curves are available, so we report their mean over tasks.
T-SAC was last evaluated at 964,500 environment steps; we report this evaluation. Table~\ref{tab:extra_baselines} summarizes all
baselines at 1M environment steps.

\begin{table}[h]
    \centering
    \caption{IQM success of SEAR and all baselines at 1M environment steps on the 20 hardest and on all 50 Metaworld tasks.}
    \label{tab:extra_baselines}
    \begin{tabular}{lcc}
        \toprule
        Method & Metaworld-Hard (20 tasks) & Metaworld (50 tasks) \\
        \midrule
        SEAR-10 & 84\% & 98\% \\
        SimbaV2 & 62\% & 91\% \\
        CQN-AS  & 61\% & -- \\
        T-SAC   & 13\% & 68\% \\
        SAC     & 14\% & 49\% \\
        TOP-ERL & --   & 36\% \\
        \bottomrule
    \end{tabular}
\end{table}

\section{Evaluation}

    \subsection{Metaworld}

    Metaworld~\cite{yu2021metaworldbenchmarkevaluationmultitask,mclean2026metaworld} is an open-source benchmark for meta-reinforcement learning and multi-task 
    robotic manipulation. It contains 50 diverse manipulation tasks that span skills such as grasping, pushing, insertion, and object placement. Unlike benchmarks focused on a narrow task distribution, Metaworld provides a broad and challenging testbed for algorithms that aim to generalize across behaviors.

\section{Computational Efficiency}\label{computational-efficiency}

As implementation details of the replay buffer and the environment
obscure the computational requirements of the methods, we compare the
time for inference (i.e., sampling a chunk) and policy updates (i.e.,
updating the critic and policy). This ensures fairness, as CQN-AS's
replay buffer implementation is optimized for image observations and
thus less efficient for state observations. All measurements were
obtained with UTD=1 on a single RTX 4500 Blackwell GPU, separately from the A100
training runs reported in Appendix~\ref{app:compute} (Table~\ref{tab:compute_timing}).

\begin{table}[h]
    \centering
    \caption{Inference time and update time of SEAR, SEAR with an MLP critic, CQN-AS and SimbaV2 for different chunk sizes, measured with UTD=1 on a single RTX 4500 Blackwell GPU. Update time per environment step accounts for how often each method updates.}
    \label{tab:compute_timing}
    \begin{tabular}{lrrr}
        \toprule
        Method & Inference [ms] & Update [ms] & Update per Env.\ Step [ms] \\
        \midrule
        CQN-AS-1 & 5.0 & 14.6 & 14.6 \\
        CQN-AS-5 & 5.5 & 18.6 & 18.6 \\
        CQN-AS-10 & 5.8 & 29.8 & 29.8 \\
        CQN-AS-20 & 6.3 & 60.4 & 60.4 \\
        \midrule
        SEAR-1 & 1.7 & 9.7 & 9.7 \\
        SEAR-5 & 1.7 & 22.6 & 9.0 \\
        SEAR-10 & 1.7 & 50.7 & 10.1 \\
        SEAR-20 & 1.7 & 126.6 & 12.7 \\
        \midrule
        SEAR-1 (MLP) & 1.7 & 5.0 & 5.0 \\
        SEAR-5 (MLP) & 1.7 & 7.0 & 2.8 \\
        SEAR-10 (MLP) & 1.7 & 7.9 & 1.6 \\
        SEAR-20 (MLP) & 1.7 & 9.0 & 0.9 \\
        \midrule
        SimbaV2 & 1.7 & 5.0 & 5.0 \\
        \bottomrule
    \end{tabular}
\end{table}

First, we observe that SEAR(-MLP) and SimbaV2 (which share the same
actor) are faster at inference than the critic-only method CQN-AS.

Second, regarding update time, we observe that all action-chunking
methods exhibit increasing update times with higher chunk sizes. As a
reference point, SimbaV2 spends only 5ms per environment step on
updates. As SEAR does only one update per chunk (which is on average
executed halfway, due to random replanning), SEAR spends 9-13ms per
environment step on updates, roughly 1.8-2.5x the time of SimbaV2.

CQN-AS does one update per environment step, which takes between 14ms
and 60ms, depending on chunk size.

SEAR-10's \textasciitilde2x update cost per environment step means
running SEAR-10 for 500k samples costs about as much update wall-clock
time as running SimbaV2 for 1M samples. On Metaworld-Hard, SimbaV2 has
an IQM success of 62\% at 1M steps, which is better than SEAR-10's 48\%
success at 500k steps. Under a comparable compute budget, SimbaV2 thus
has an advantage. How relevant this is when including time spent on
environment interaction, evaluation and buffer operations will be
application dependent.

Future work could improve the runtime of SEAR in multiple ways. The most
straightforward way would be to exchange the transformer for a more
compute-efficient backbone like Mamba~3~\cite{lahoti2026mamba}. Furthermore, the critic
update could be made more compute-efficient by using a value function
for the multi-horizon targets, like TOP-ERL.

\section{Extended Results}
\label{app:extended_results}
\begin{figure*}[t]
\centering
    \hfill
    \begin{subfigure}[b]{.39\textwidth}
        \centering
\includegraphics[width=\linewidth]{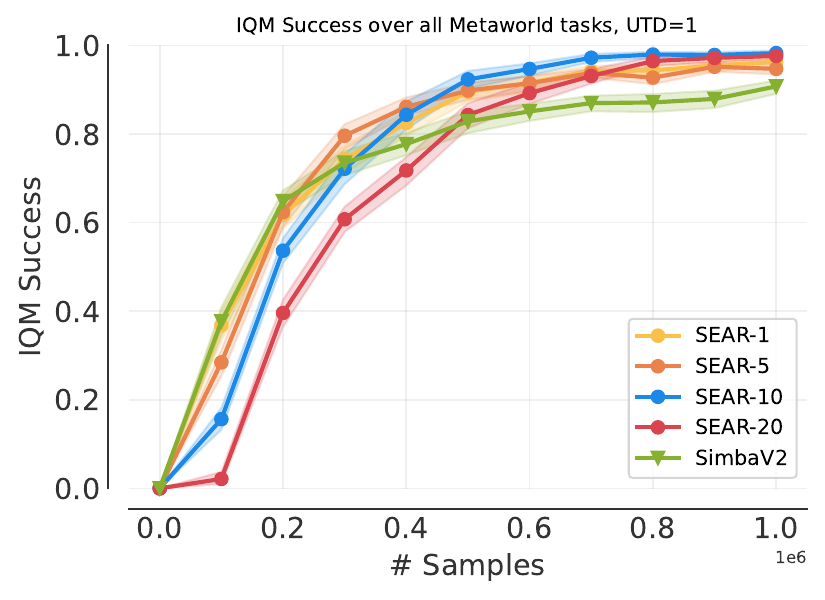}
        \caption{Sample Efficiency}
    \end{subfigure}
    \hfill
    \begin{subfigure}[b]{.59\textwidth}
        \centering
    \includegraphics[width=\linewidth]{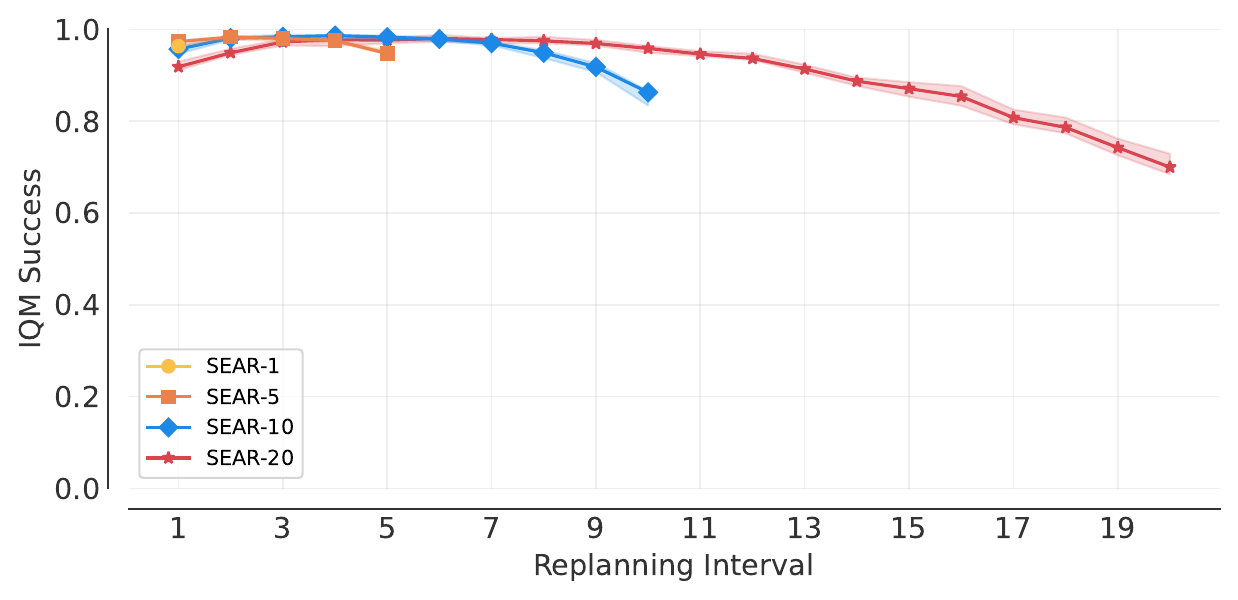}
        \caption{Effect of receding horizons }
    \end{subfigure}
    \hfill
    \caption{The experiments of Figure~\ref{fig:ablations-chunksize}, but over all 50 Metaworld environments, not just the 20 hardest that were used in the main text. SEAR still reaches a higher success rate than SimbaV2. The effect of a receding horizon is very similar to the result of the main text: $k=1$ is suboptimal, as is $k=N$. The optimal value grows with the chunk size: $k=2$ for SEAR-5, $k=4$ for SEAR-10 and $k=6$ for SEAR-20.
    }
\label{fig:ablations-chunksize-allmetaworld}
\end{figure*}

\begin{figure}[h]
\centering
\includegraphics[width=.18\linewidth]{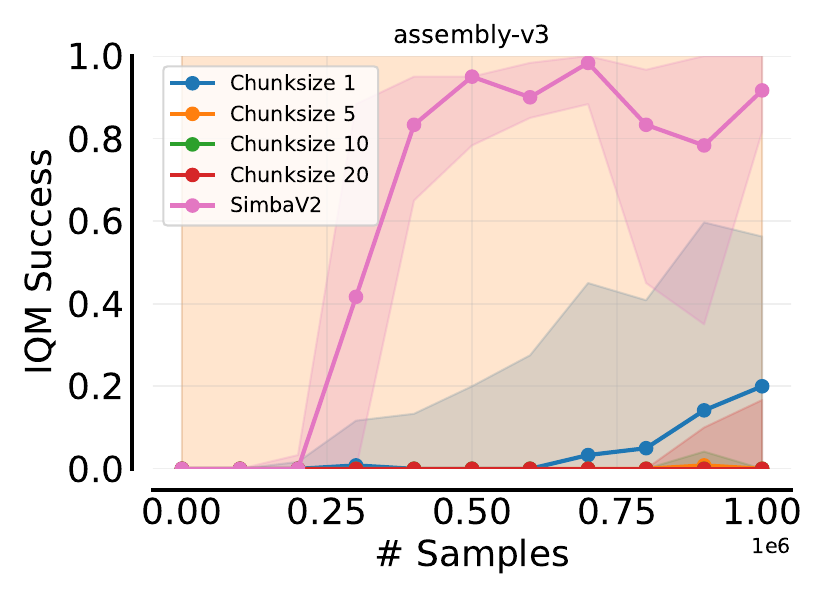}
\includegraphics[width=.18\linewidth]{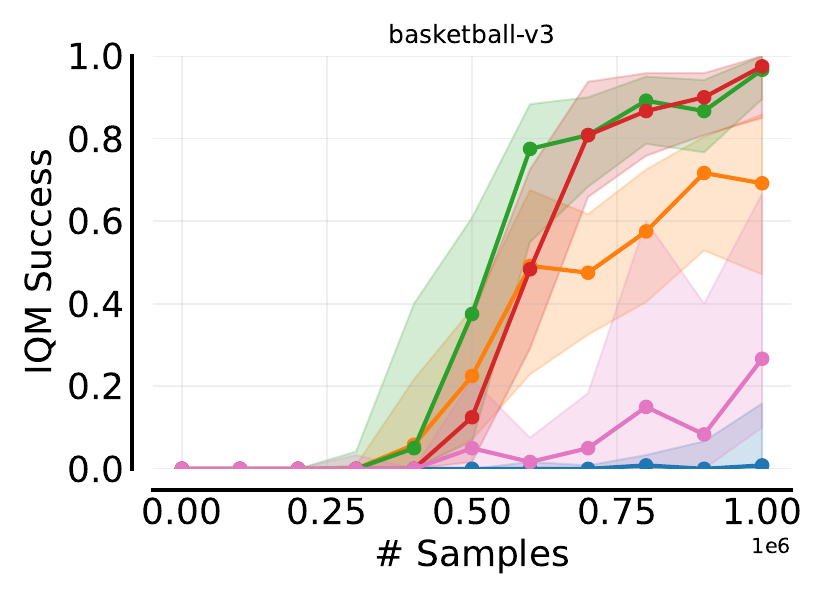}
\includegraphics[width=.18\linewidth]{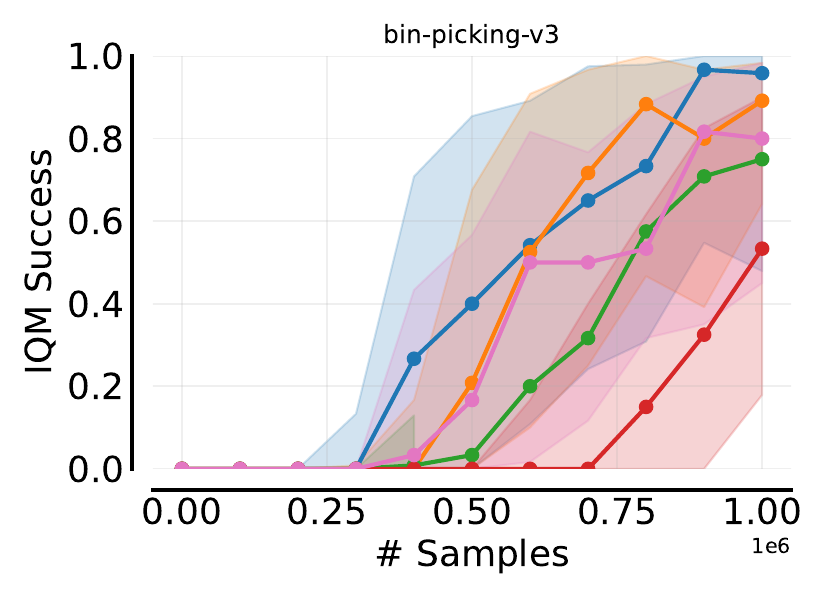}
\includegraphics[width=.18\linewidth]{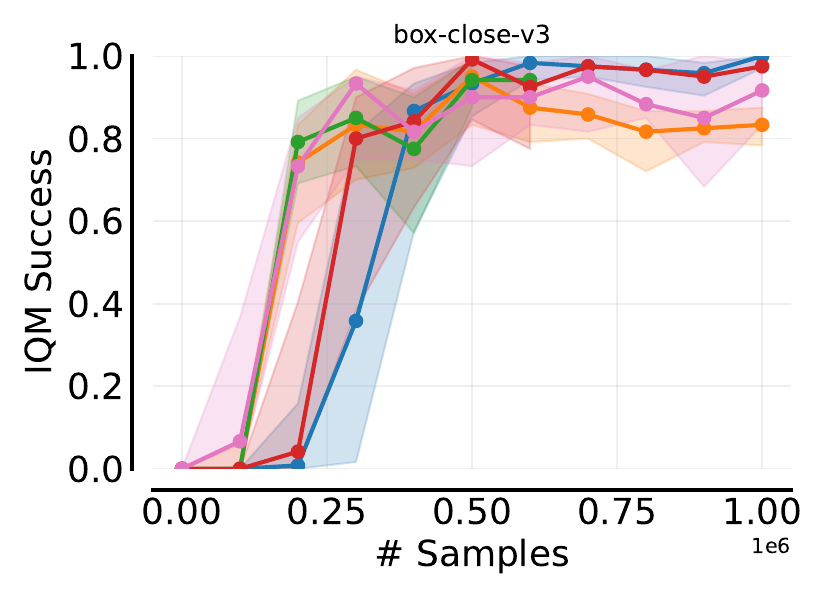}
\includegraphics[width=.18\linewidth]{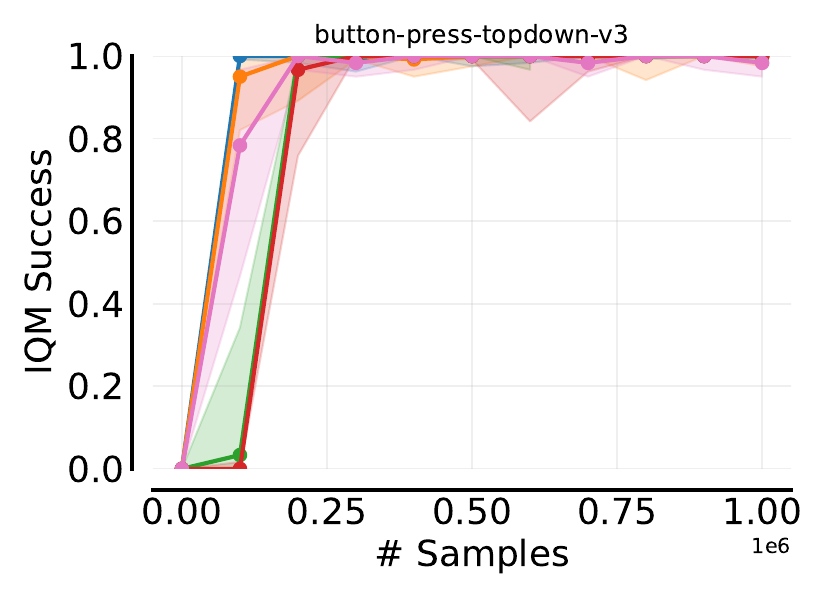} \\
\includegraphics[width=.18\linewidth]{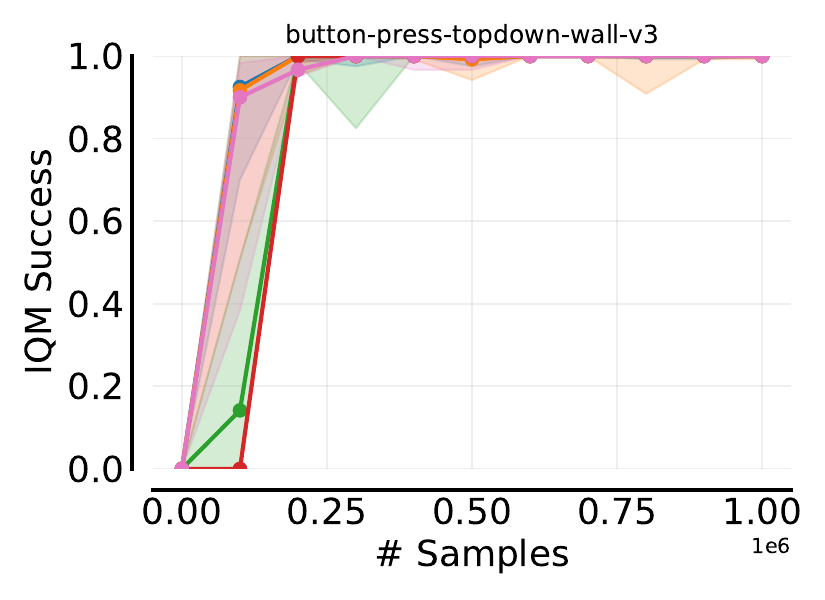}
\includegraphics[width=.18\linewidth]{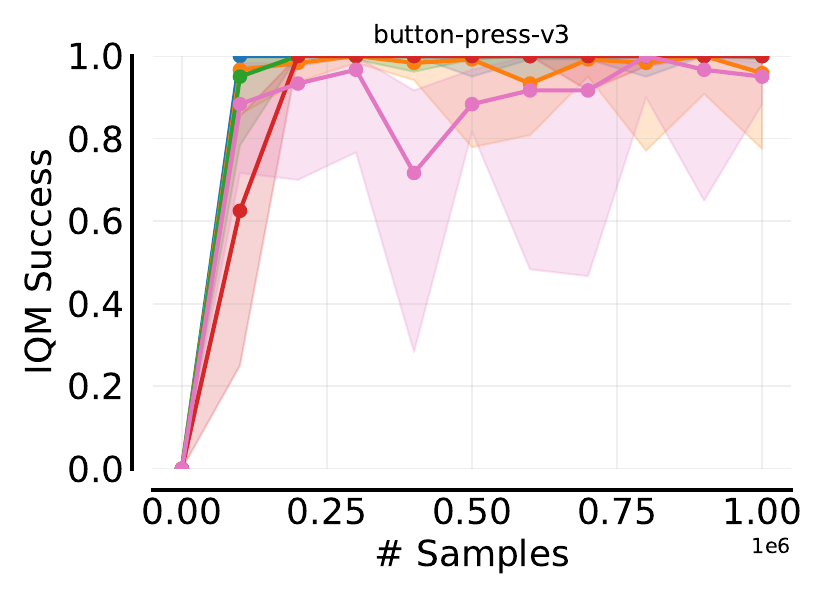}
\includegraphics[width=.18\linewidth]{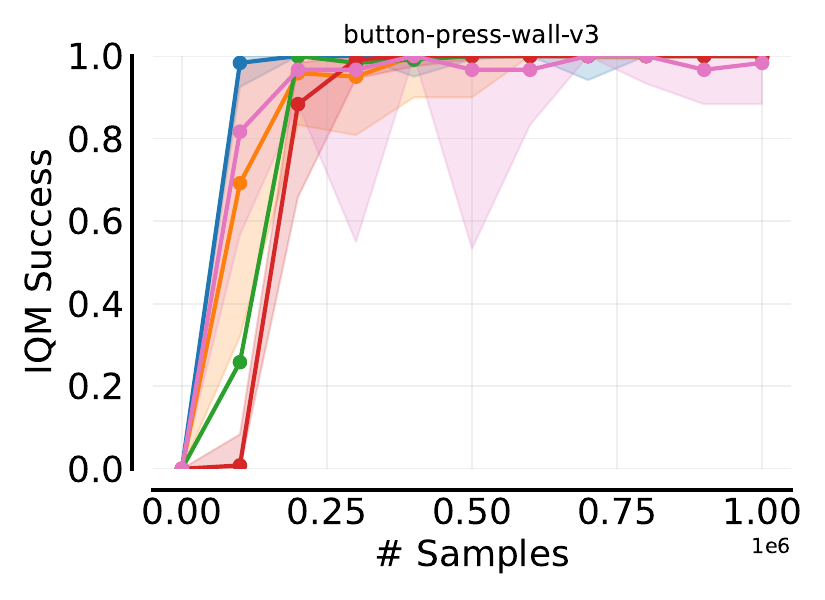}
\includegraphics[width=.18\linewidth]{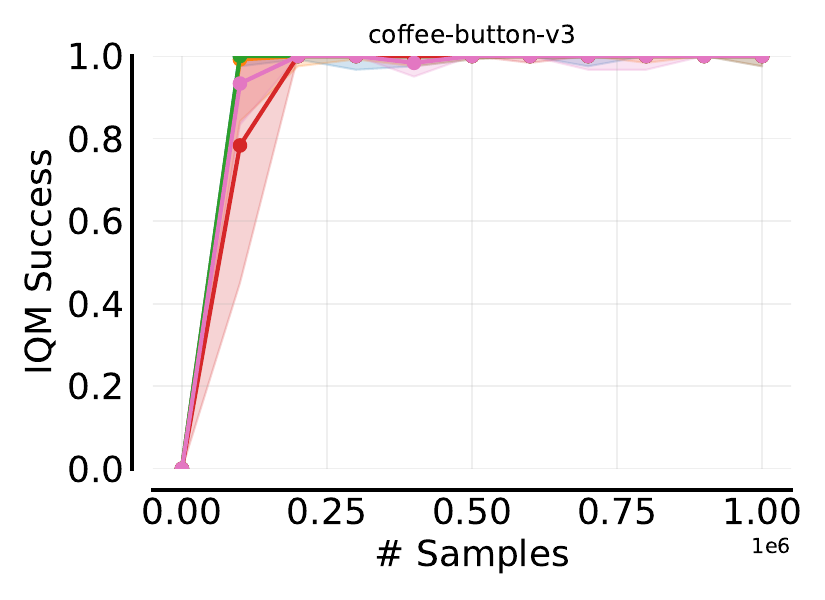}
\includegraphics[width=.18\linewidth]{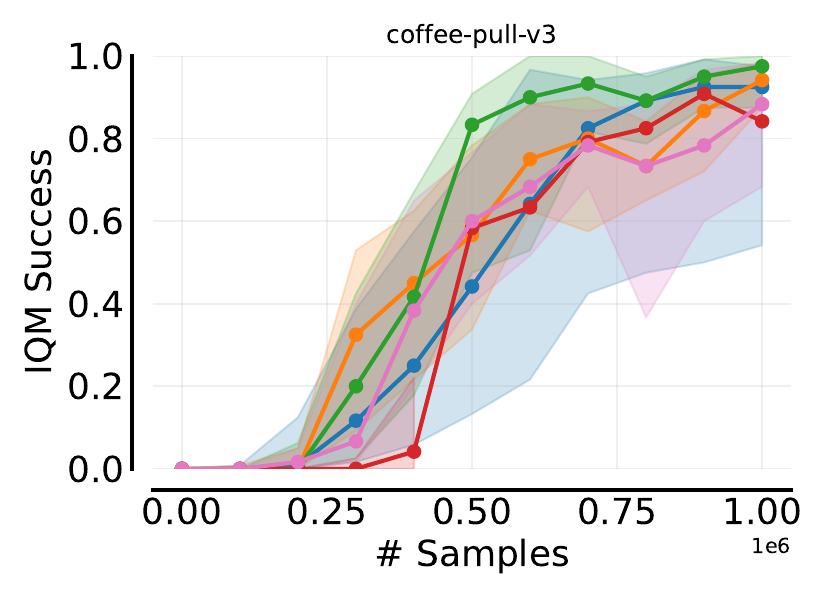} \\
\includegraphics[width=.18\linewidth]{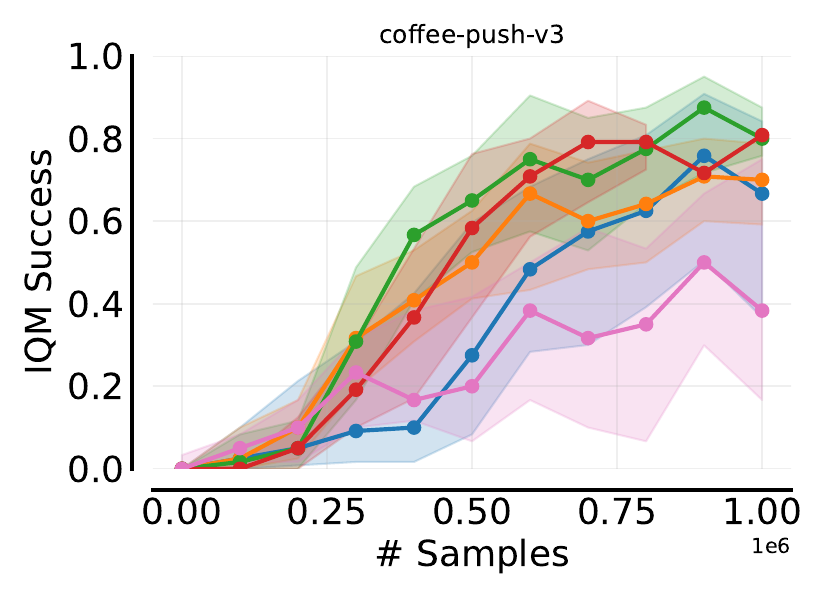}
\includegraphics[width=.18\linewidth]{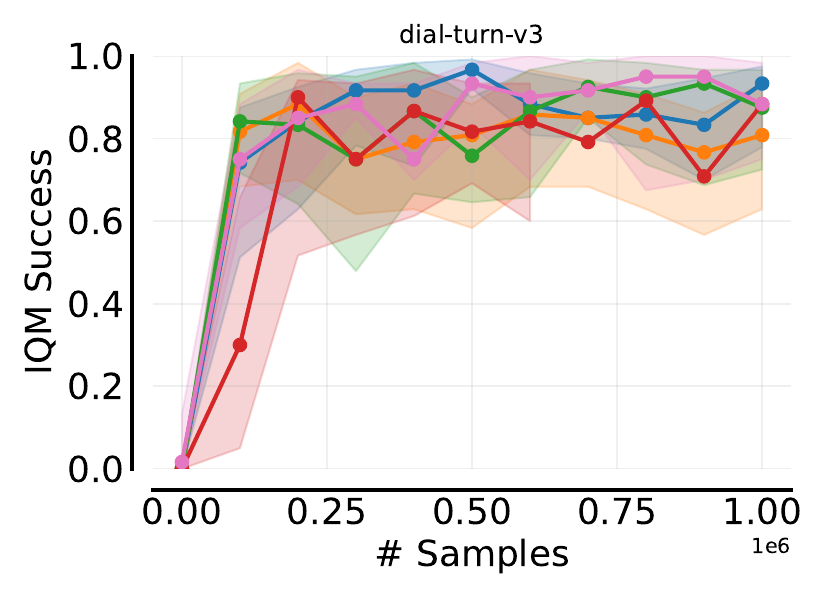}
\includegraphics[width=.18\linewidth]{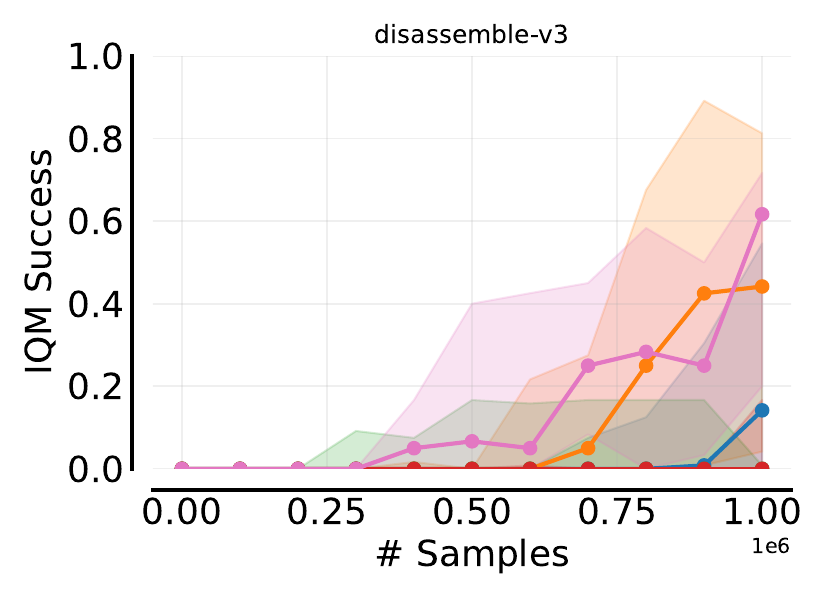}
\includegraphics[width=.18\linewidth]{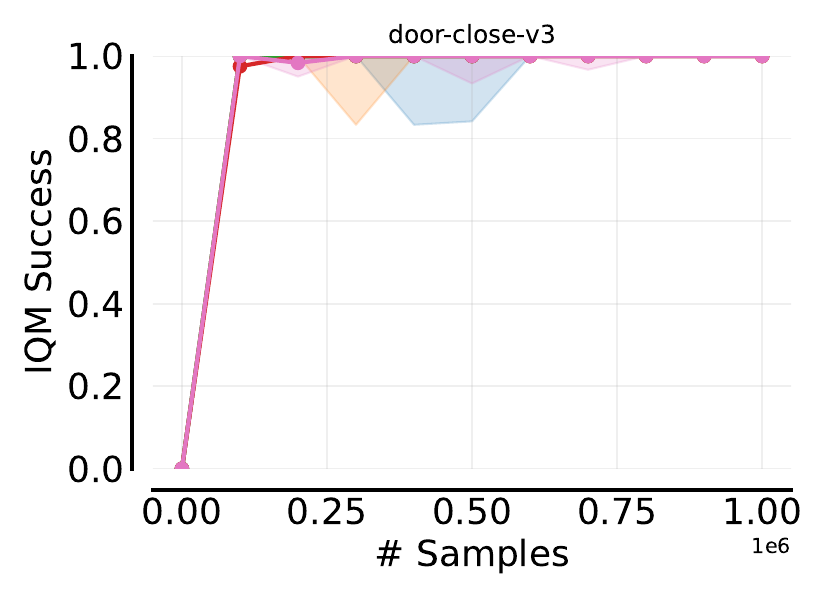}
\includegraphics[width=.18\linewidth]{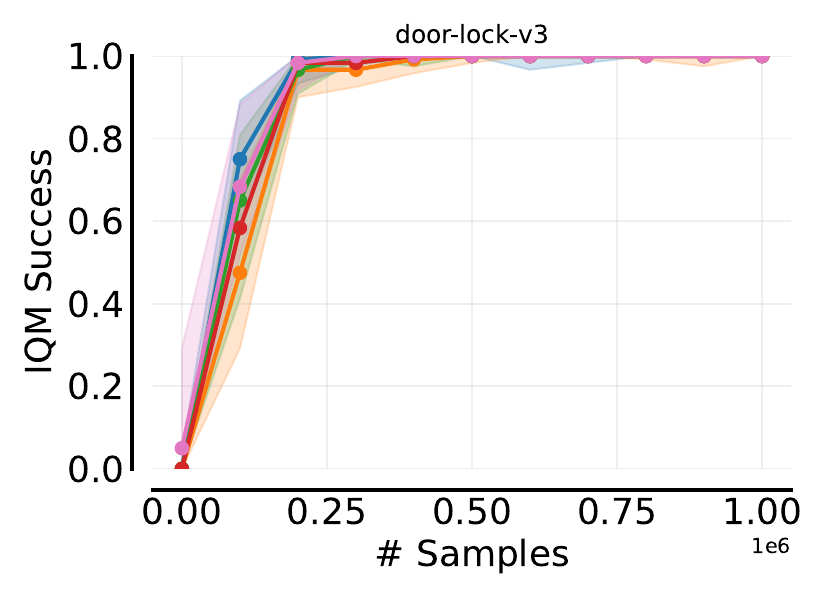} \\
\includegraphics[width=.18\linewidth]{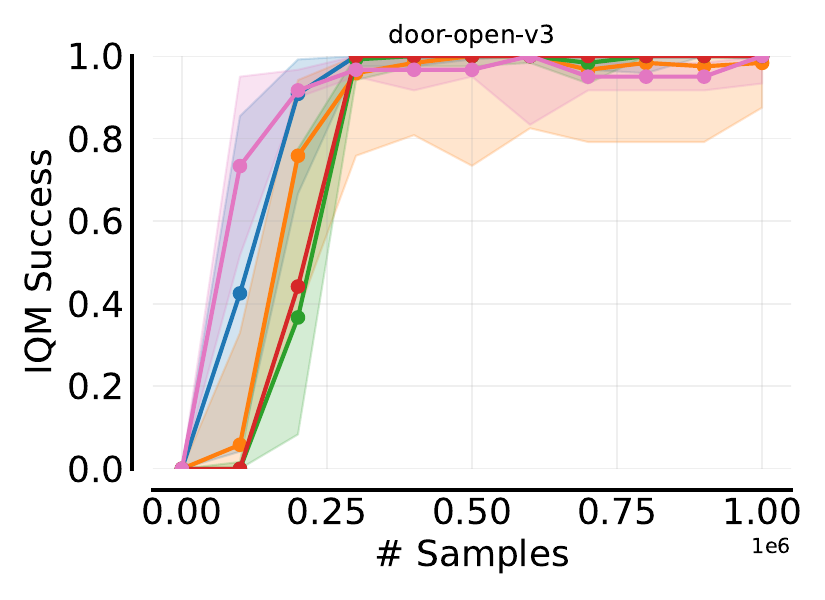}
\includegraphics[width=.18\linewidth]{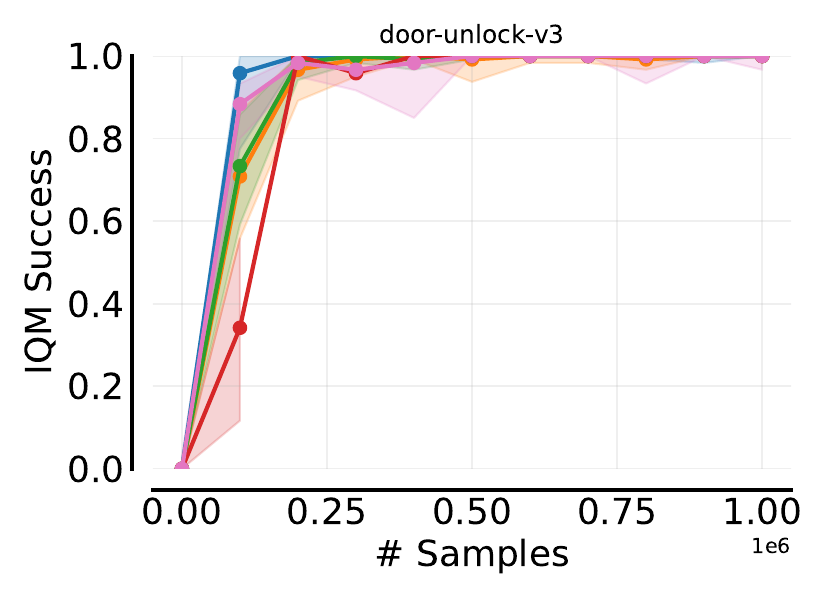}
\includegraphics[width=.18\linewidth]{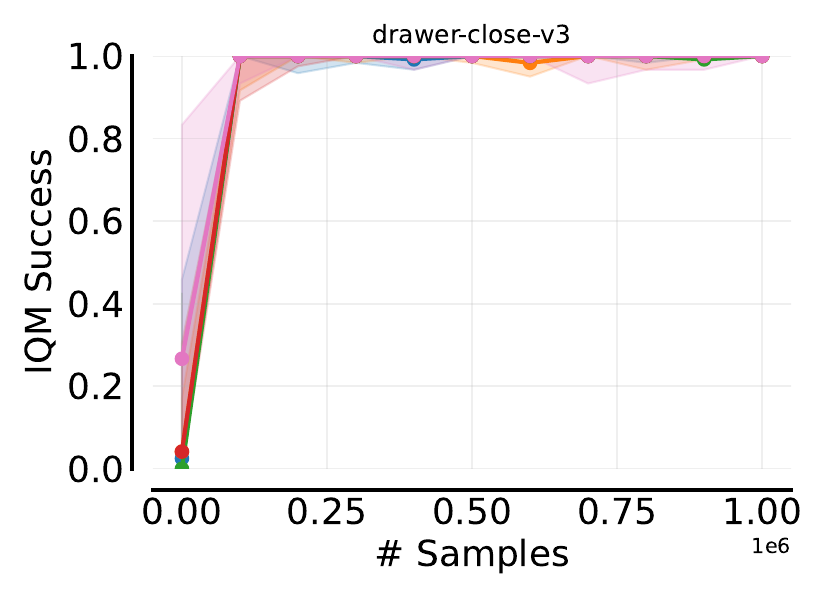}
\includegraphics[width=.18\linewidth]{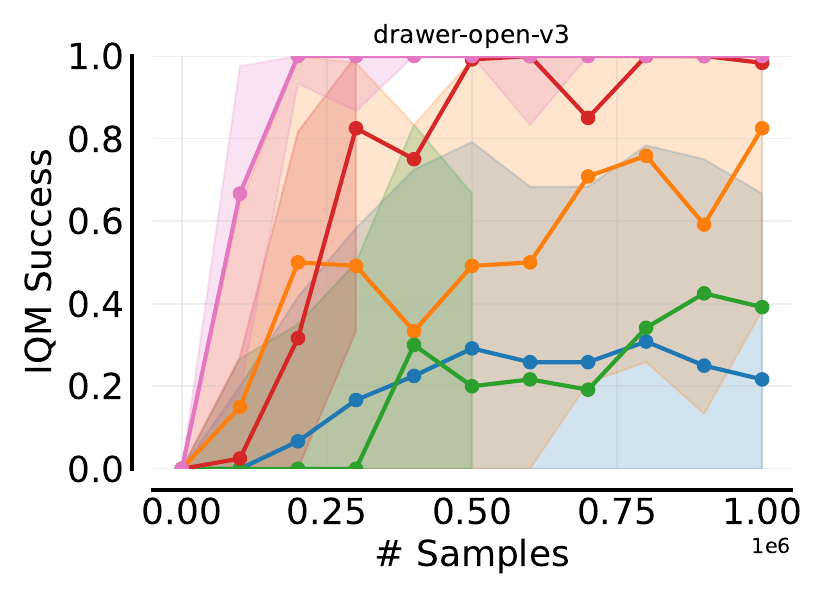}
\includegraphics[width=.18\linewidth]{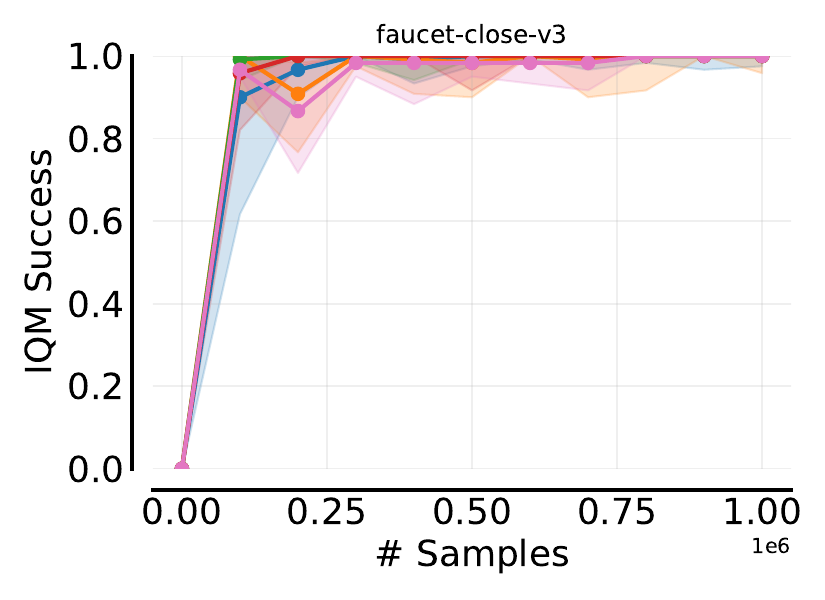} \\
\includegraphics[width=.18\linewidth]{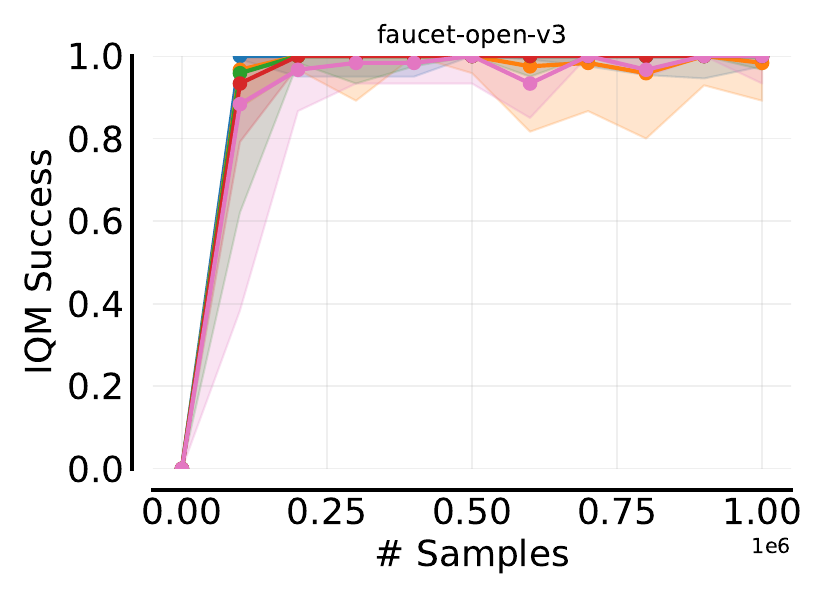}
\includegraphics[width=.18\linewidth]{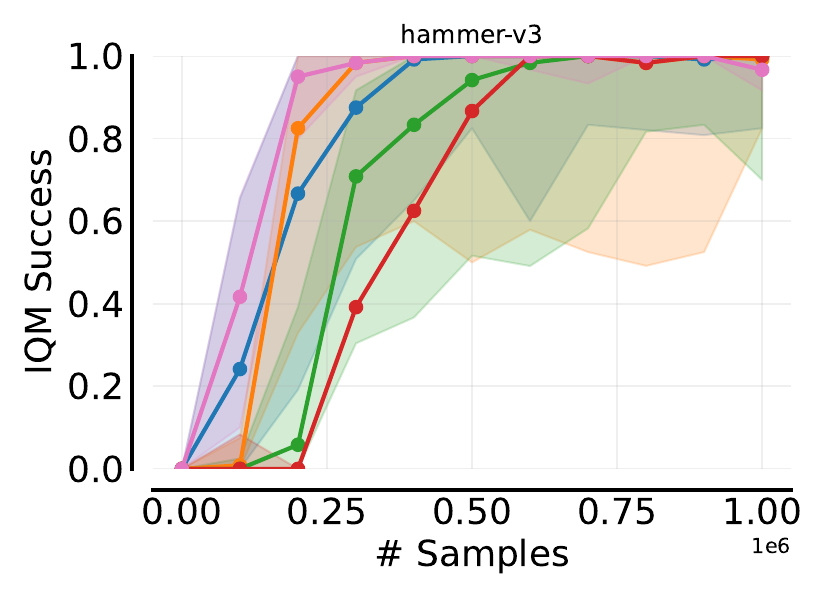}
\includegraphics[width=.18\linewidth]{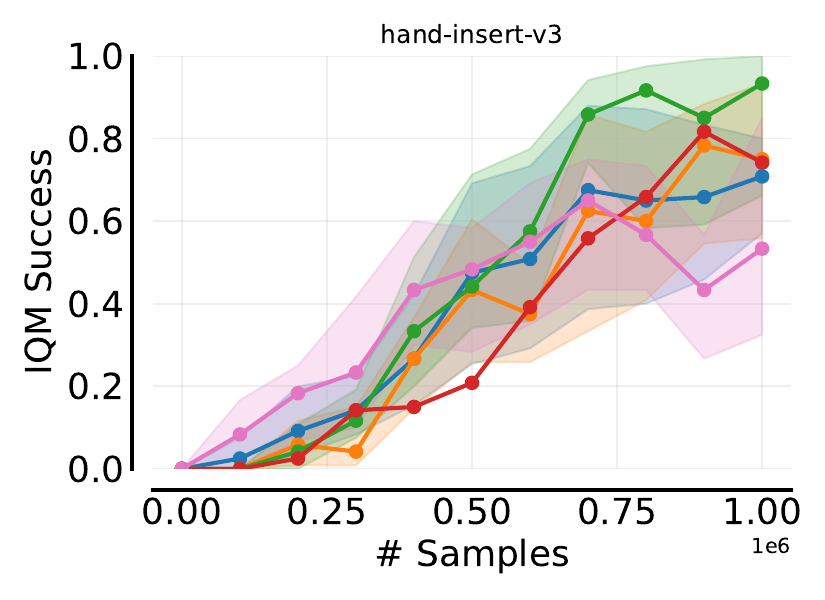}
\includegraphics[width=.18\linewidth]{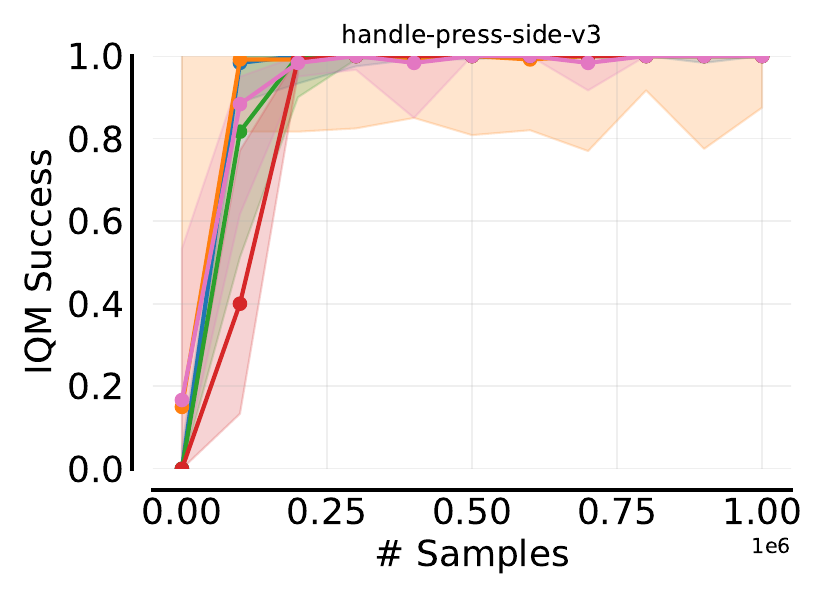}
\includegraphics[width=.18\linewidth]{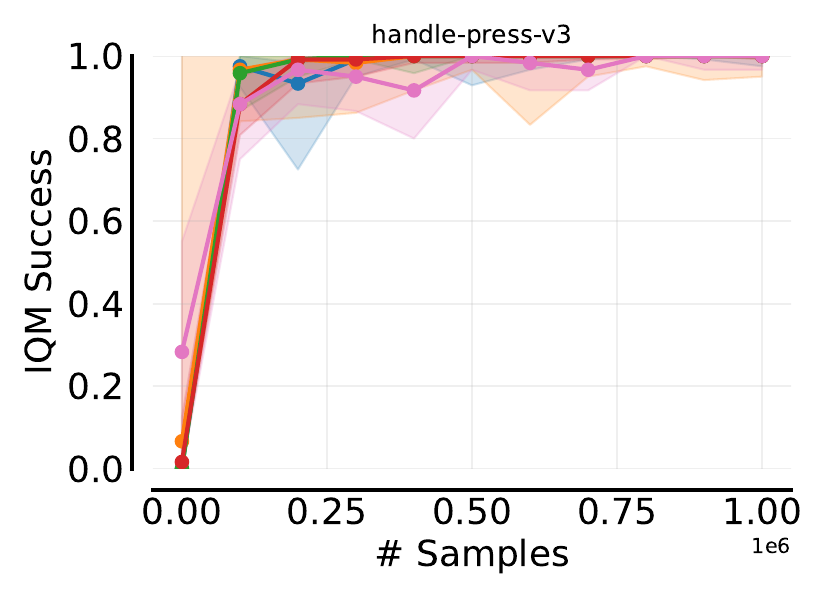} \\
\includegraphics[width=.18\linewidth]{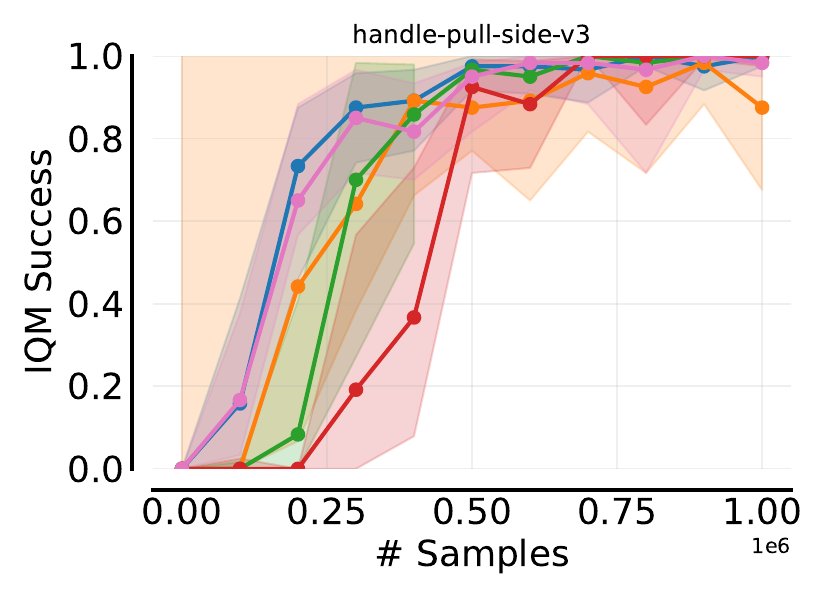}
\includegraphics[width=.18\linewidth]{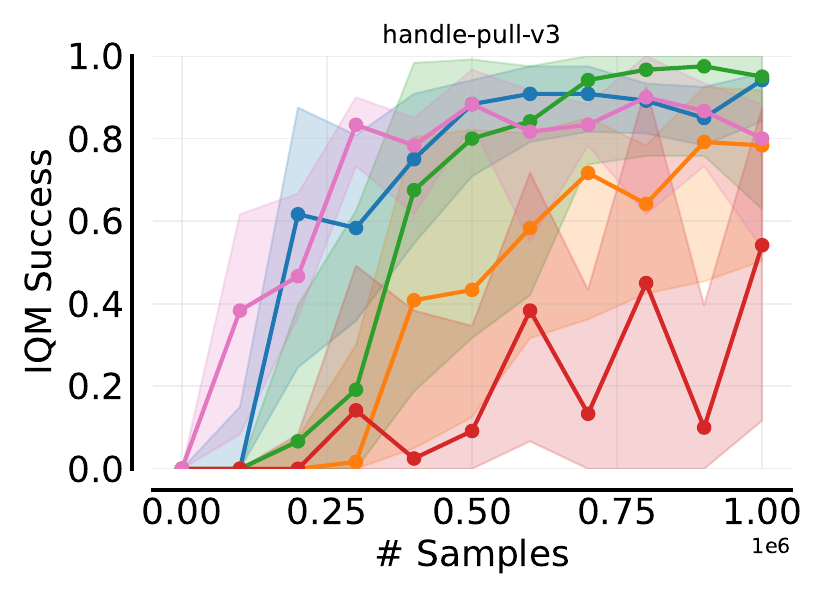}
\includegraphics[width=.18\linewidth]{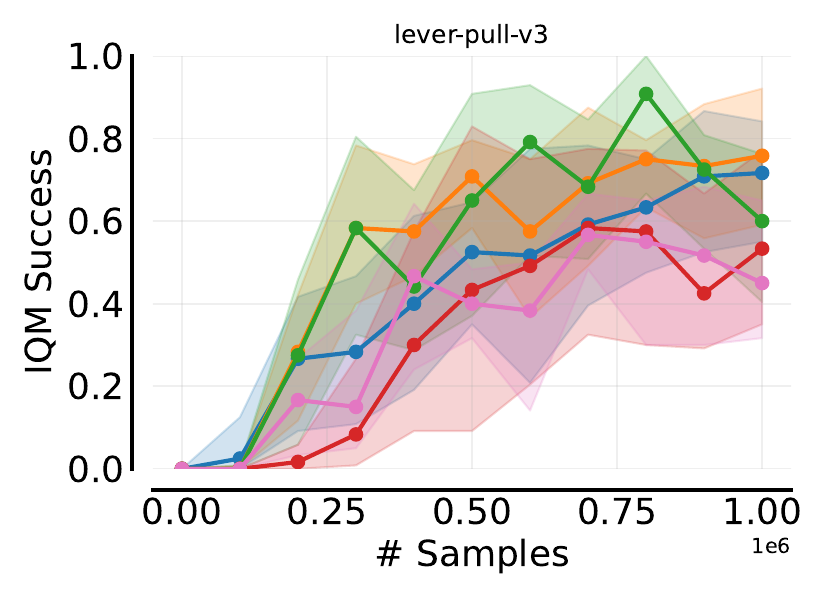}
\includegraphics[width=.18\linewidth]{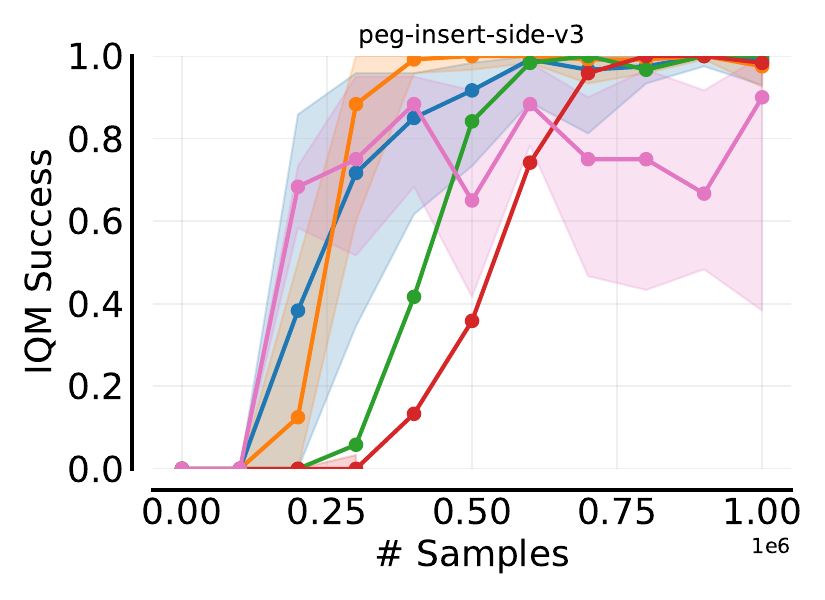}
\includegraphics[width=.18\linewidth]{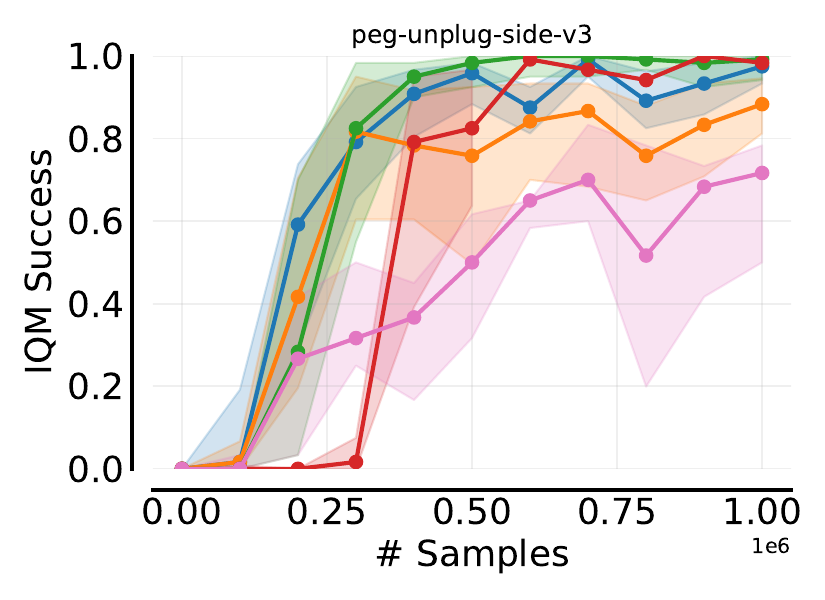} \\
\includegraphics[width=.18\linewidth]{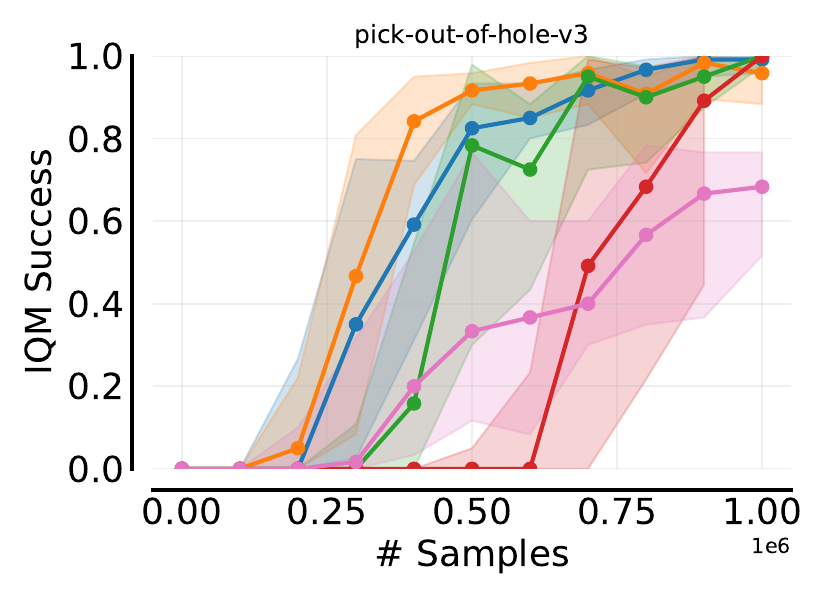}
\includegraphics[width=.18\linewidth]{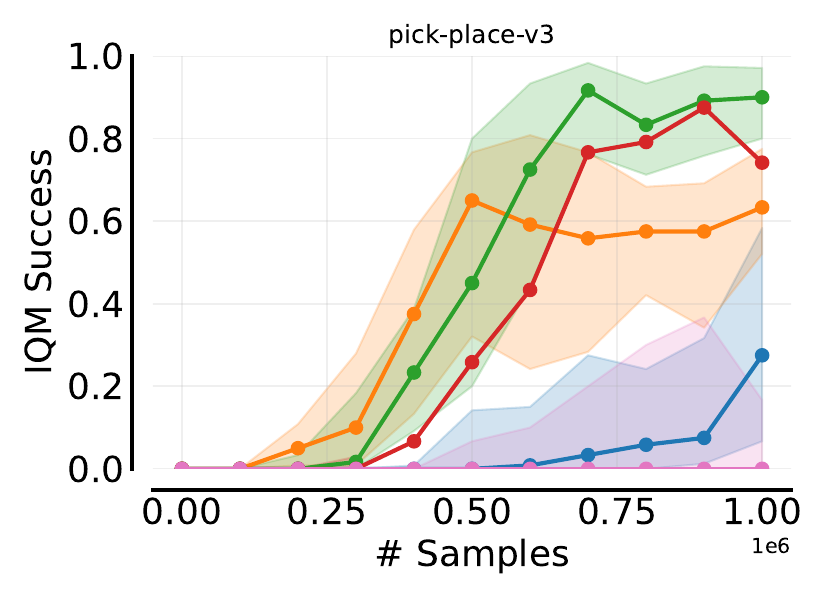}
\includegraphics[width=.18\linewidth]{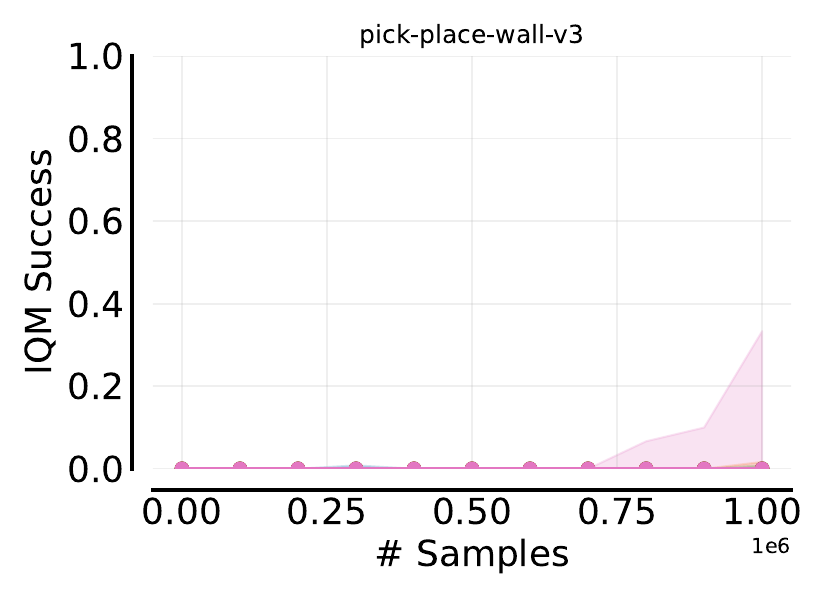}
\includegraphics[width=.18\linewidth]{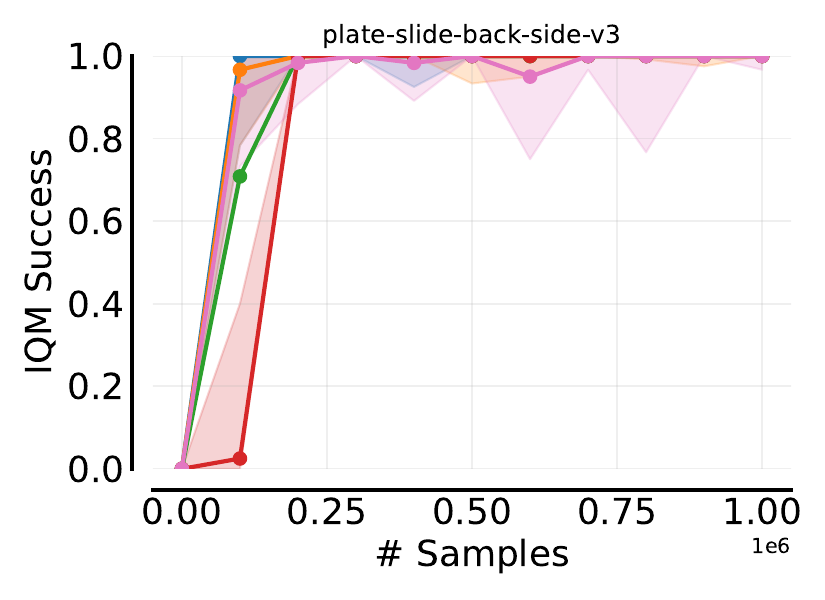}
\includegraphics[width=.18\linewidth]{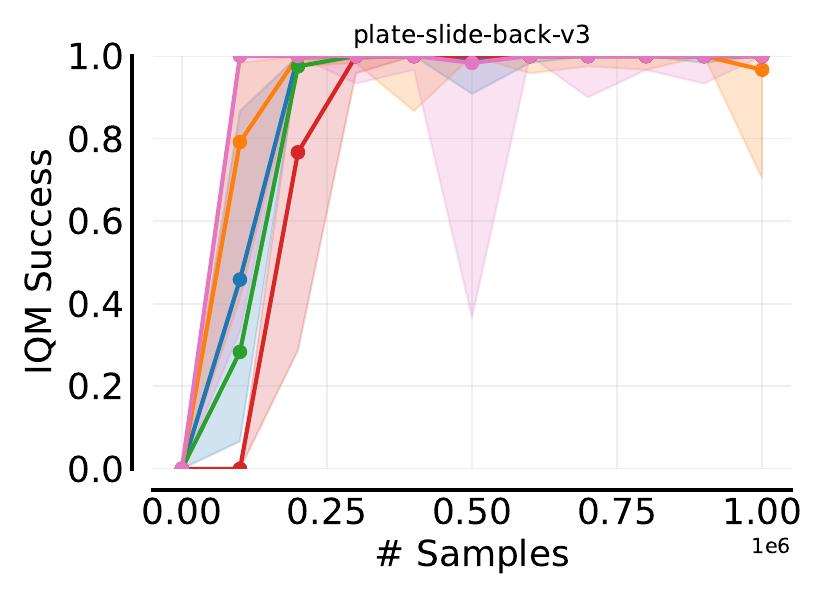} \\
\includegraphics[width=.18\linewidth]{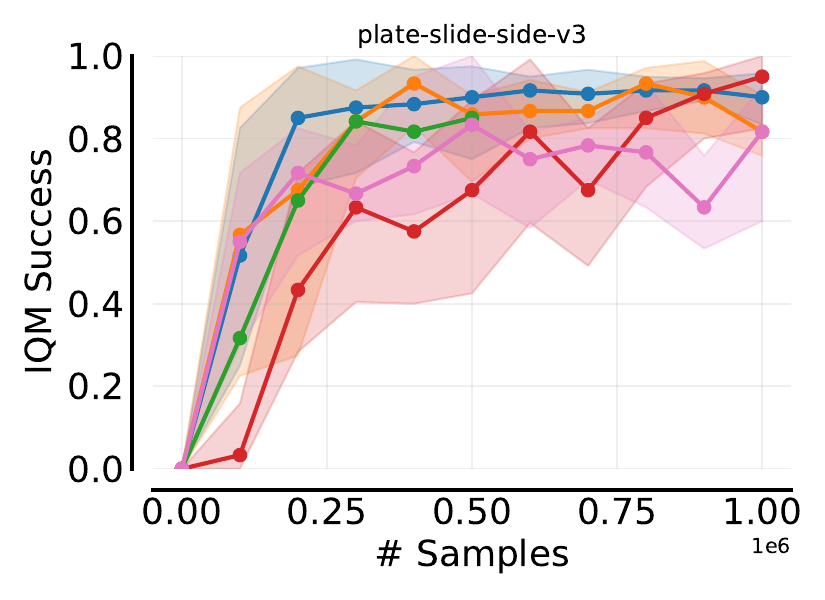}
\includegraphics[width=.18\linewidth]{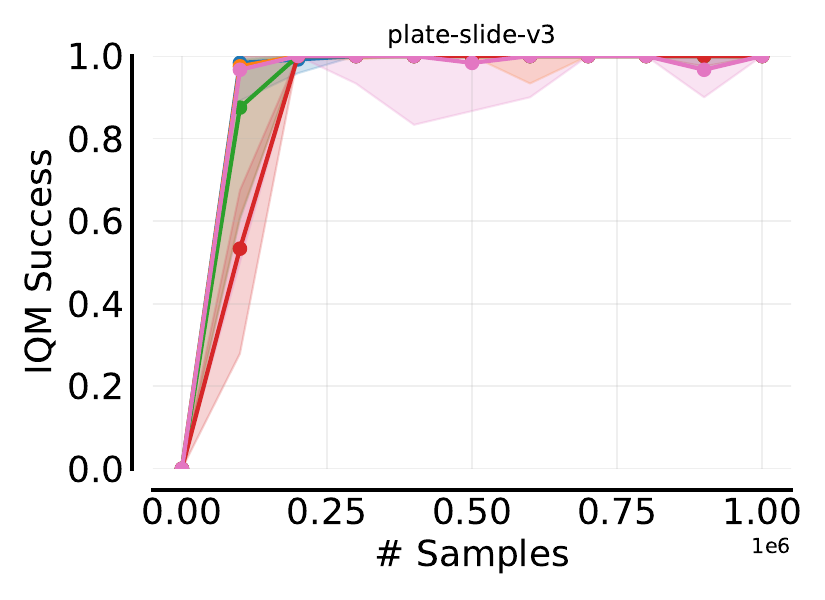}
\includegraphics[width=.18\linewidth]{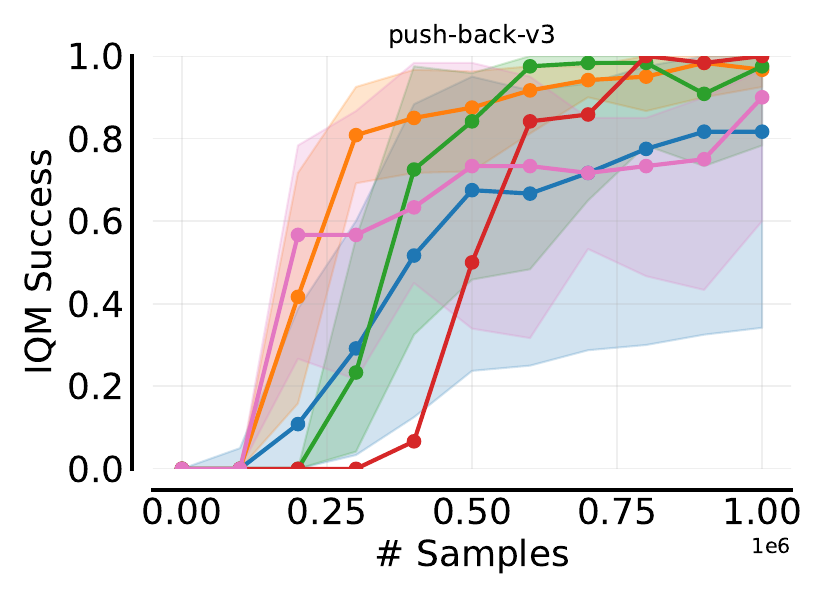}
\includegraphics[width=.18\linewidth]{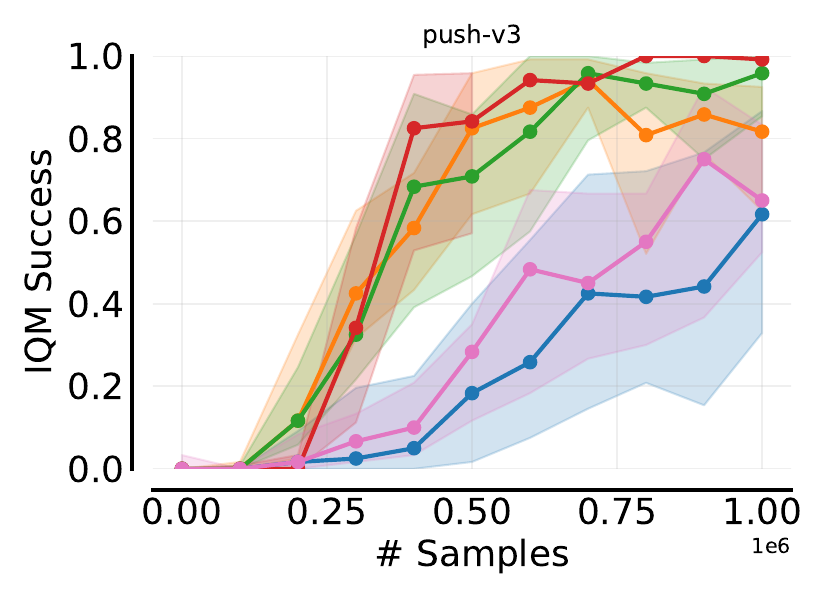}
\includegraphics[width=.18\linewidth]{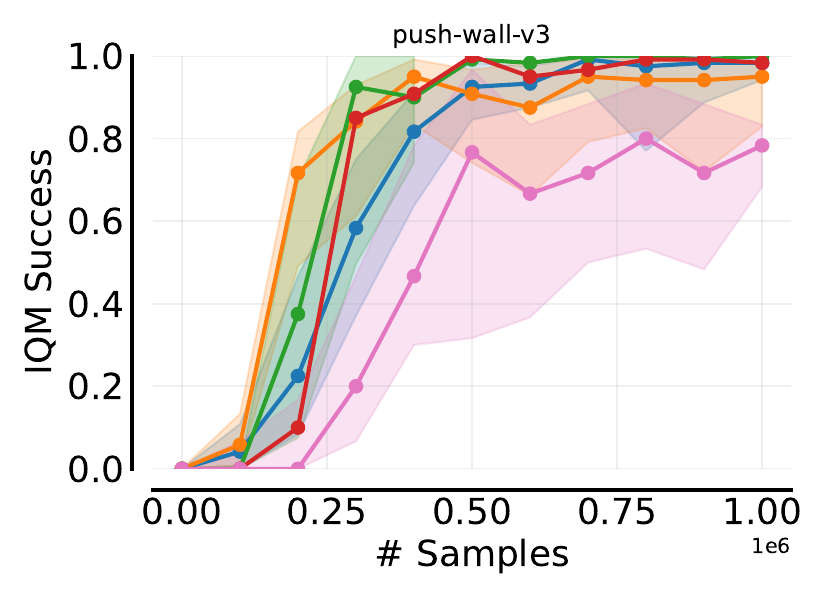} \\
\includegraphics[width=.18\linewidth]{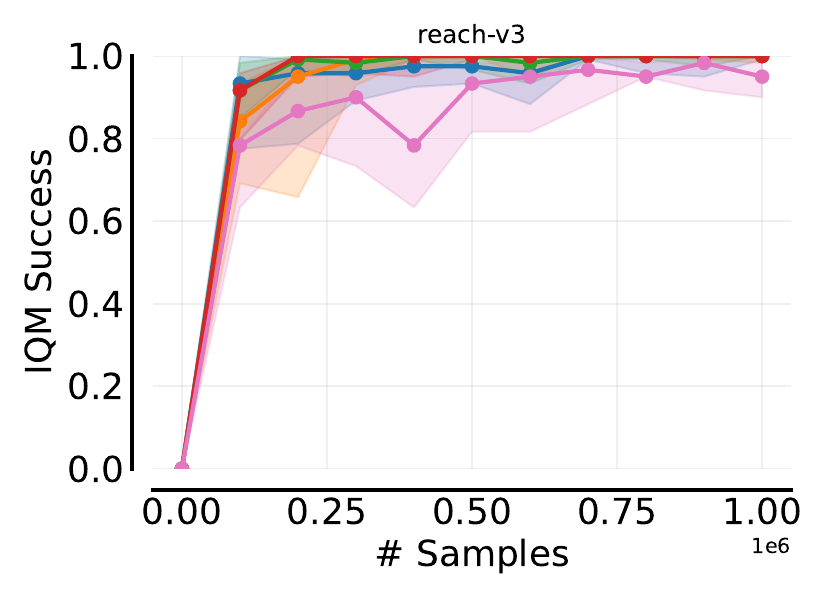}
\includegraphics[width=.18\linewidth]{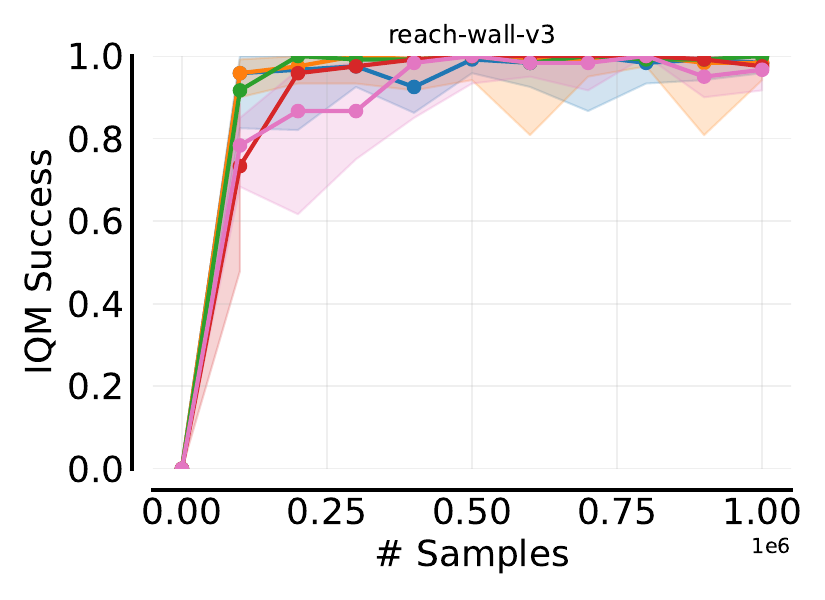}
\includegraphics[width=.18\linewidth]{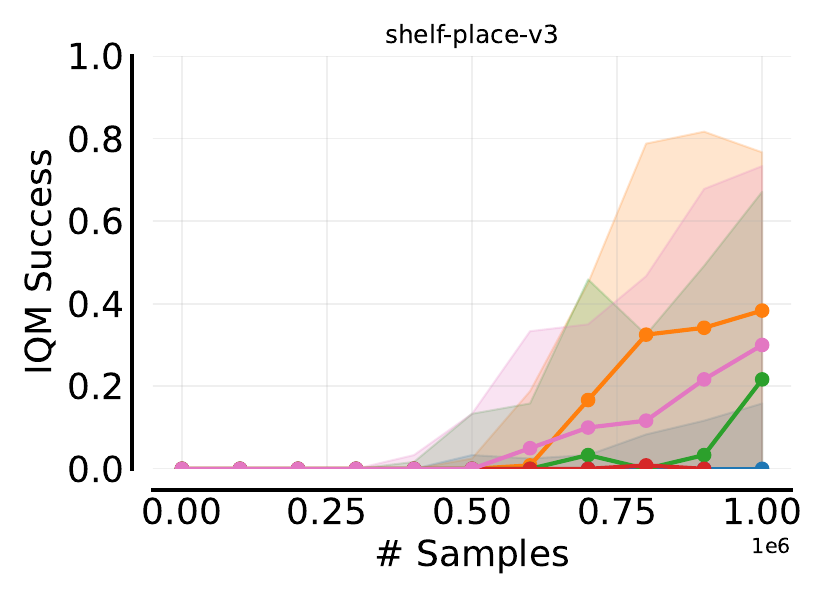}
\includegraphics[width=.18\linewidth]{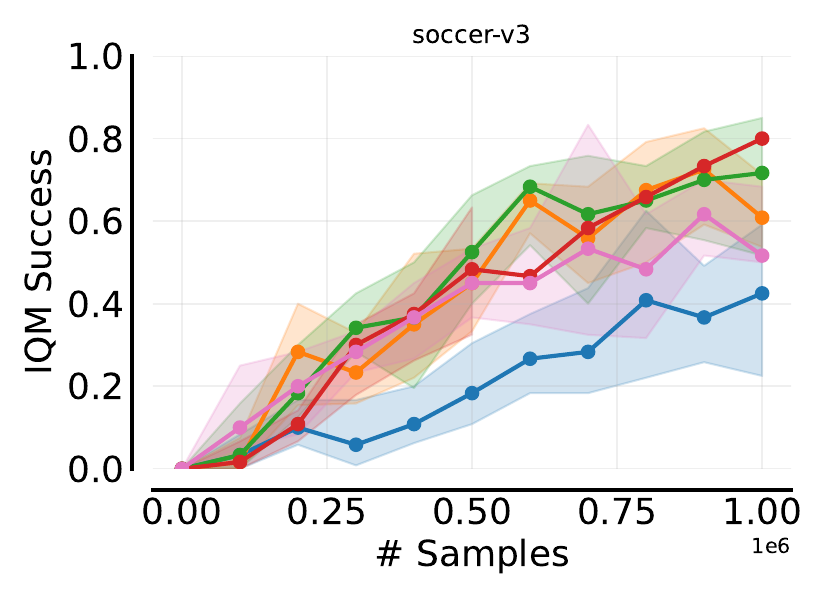}
\includegraphics[width=.18\linewidth]{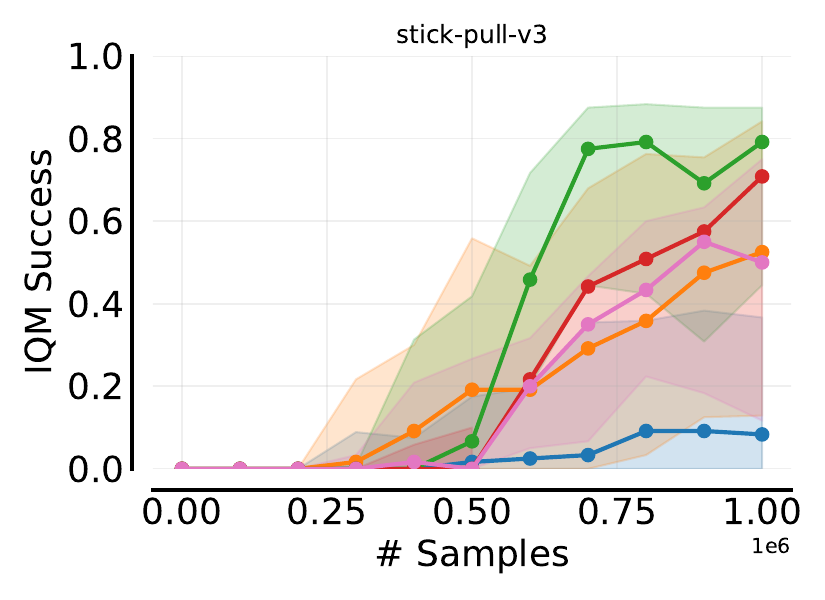} \\
\includegraphics[width=.18\linewidth]{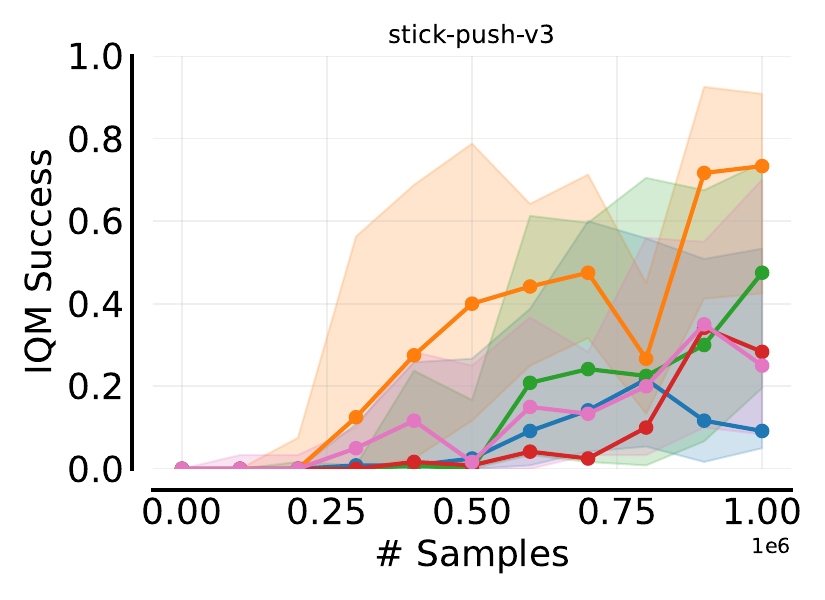}
\includegraphics[width=.18\linewidth]{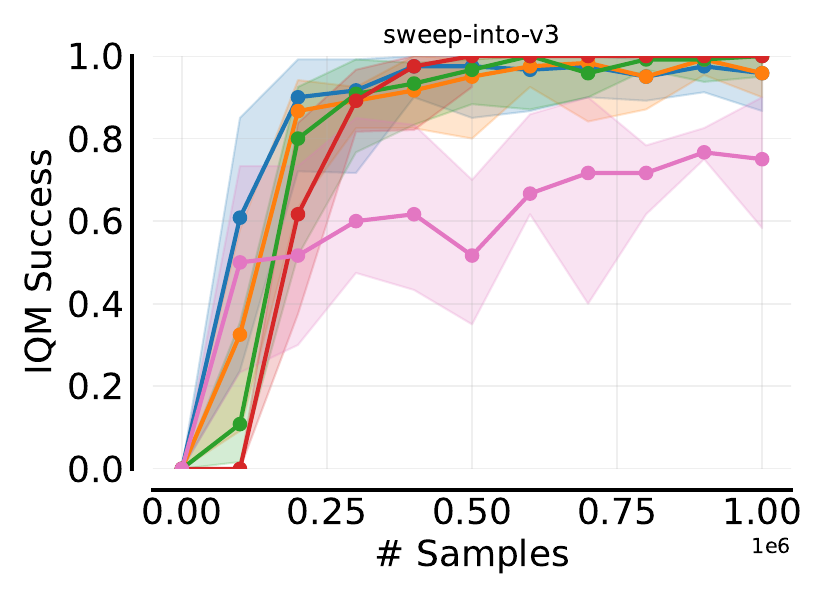}
\includegraphics[width=.18\linewidth]{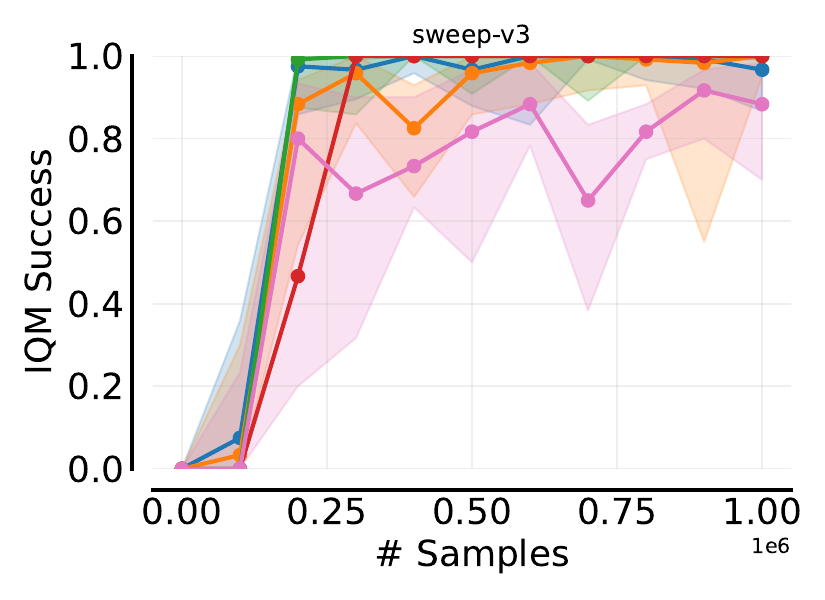}
\includegraphics[width=.18\linewidth]{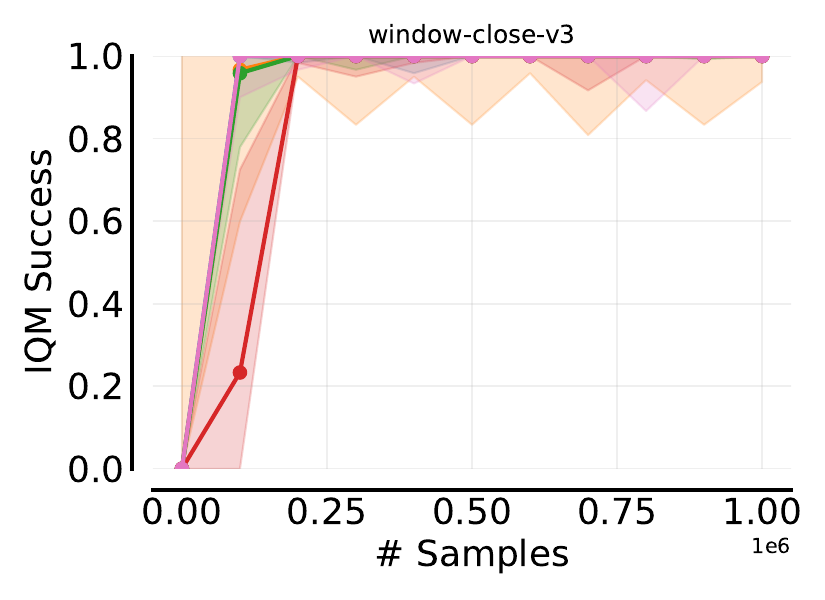}
\includegraphics[width=.18\linewidth]{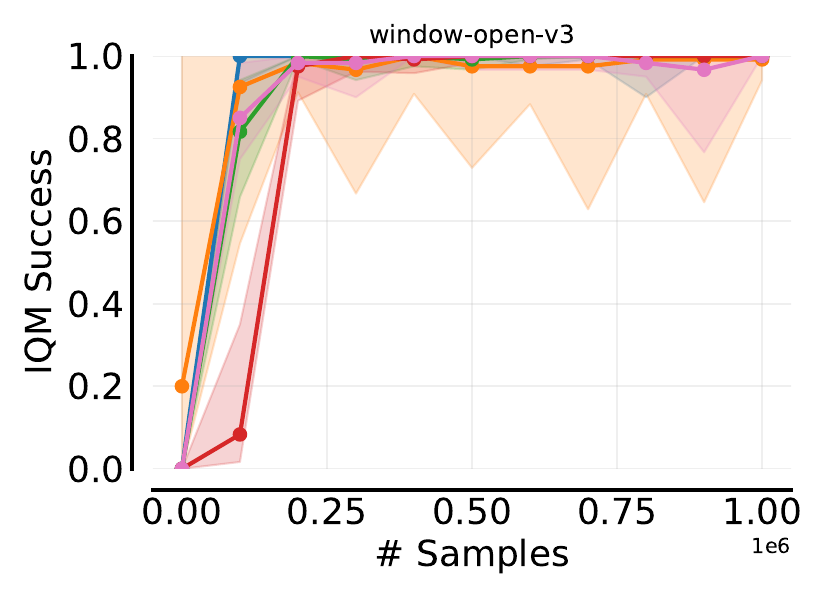}
\caption{Learning curves for all 50 Metaworld Environments.}
\label{fig:grid}
\end{figure}

\begin{figure}[h]
\centering
\includegraphics[width=.16\linewidth]{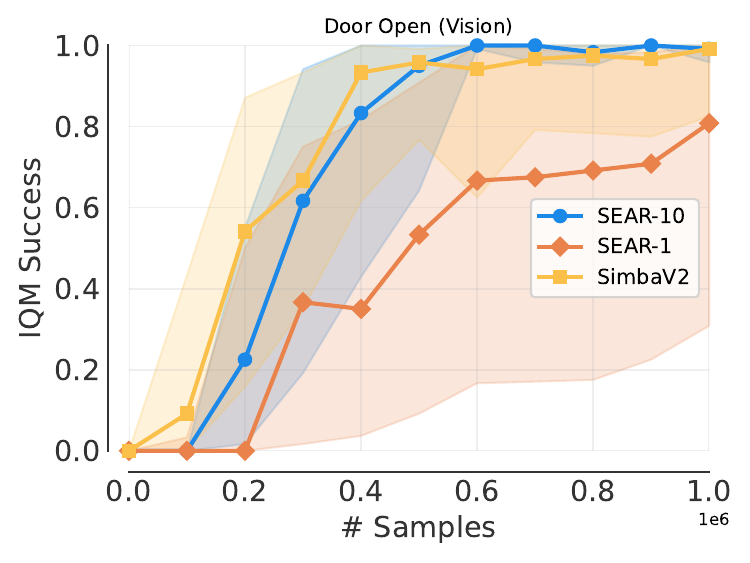}
\includegraphics[width=.16\linewidth]{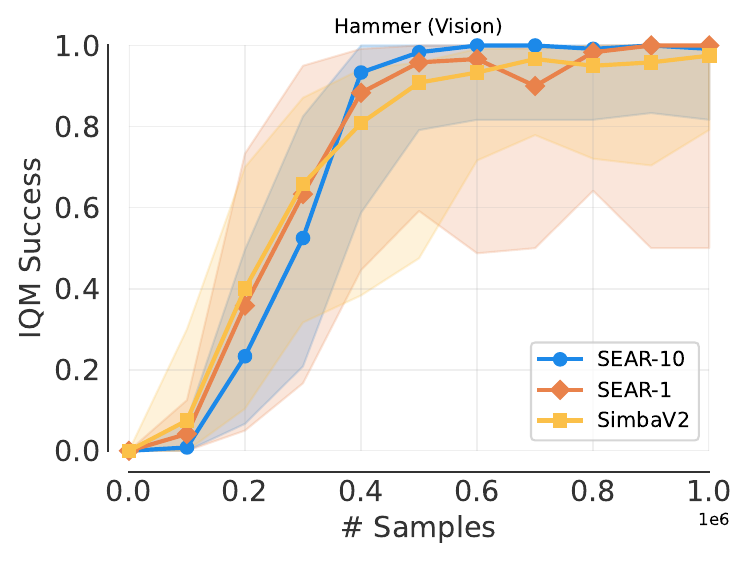}
\includegraphics[width=.16\linewidth]{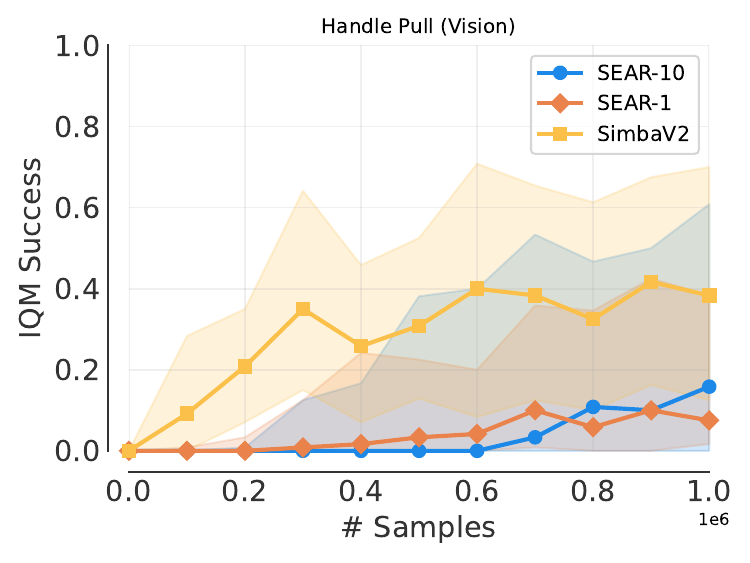}
\includegraphics[width=.16\linewidth]{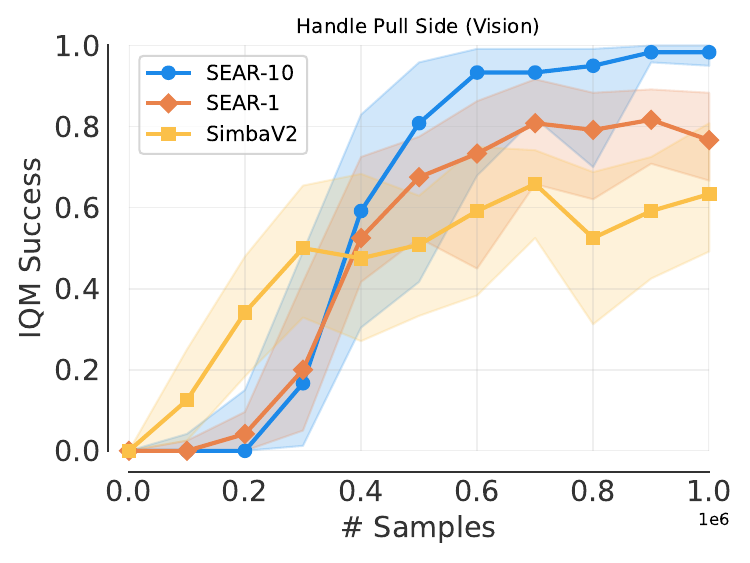}
\includegraphics[width=.16\linewidth]{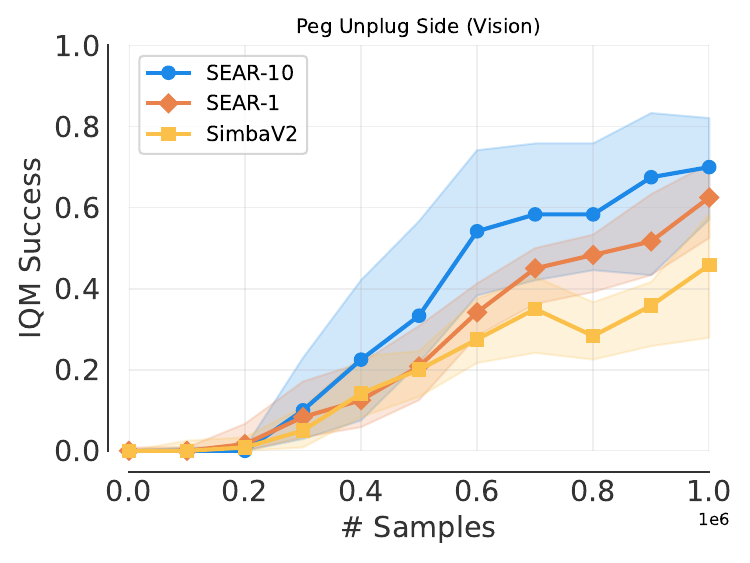}
\includegraphics[width=.16\linewidth]{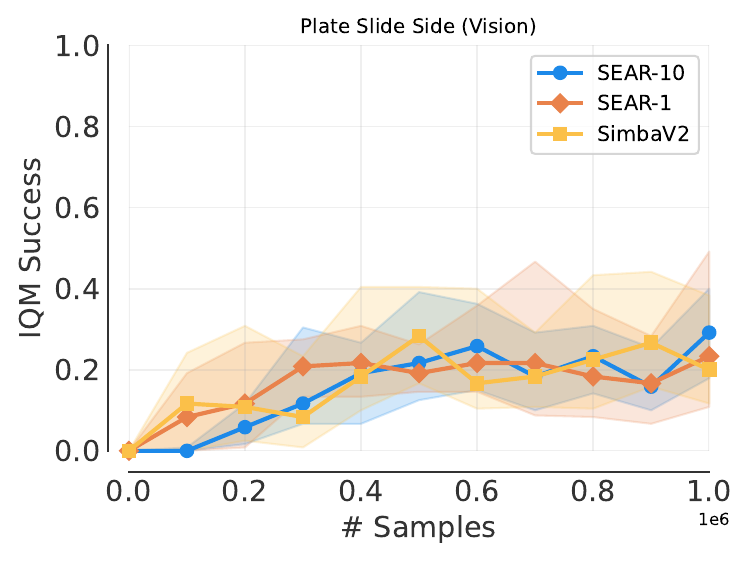} \\
\caption{Learning curves for the 6 Metaworld Environments used in the visual RL experiments.}
\label{fig:visual_grid}
\end{figure}

As Metaworld contains a number of very simple tasks, we focus our analysis in the main text on a set of the
20 hardest tasks, selected based on SimbaV2's performance in preliminary experiments. In this section we provide the per-environment performance of the methods discussed in the main text. Table~\ref{tab:metaworld-20-baselines-table} provides a list of the 20 hard tasks used in the main text and the per-task performance of SEAR-10, CQN-AS and SimbaV2. Table~\ref{tab:metaworld-easy-table} contains the tasks that were not used for the main text along with the per-task performance of SEAR-{1,5,10,20} and SimbaV2. To demonstrate that the key results of the main text, namely SEAR's superior sample efficiency and its effectiveness when using receding horizons are not 
dependent on the choice of tasks, we reproduced Figure~\ref{fig:ablations-chunksize} of the main text over all
50 tasks in Figure~\ref{fig:ablations-chunksize-allmetaworld}. Finally, we include learning curves for all environments in Figure~\ref{fig:grid}.

\definecolor{bestcolor}{HTML}{B4C7E7}
\definecolor{overlapcolor}{HTML}{D9E1F2}
\definecolor{cigray}{HTML}{999999}
\newcommand{\ci}[1]{\textcolor{cigray}{\footnotesize$\pm #1$}}
\newcommand{\asymci}[2]{\footnotesize%
    $\textcolor{cigray}{%
        \begin{smallmatrix} 
          +#1 \\ -#2 
        \end{smallmatrix}%
    }$
}

\begin{table}[]
    \centering
\begin{tabular}{Sl | c c c c c c}
     & SEAR-1 & SEAR-5 & SEAR-10 & SEAR-20 \\
    \midrule
    Average & $0.57$ \asymci{0.07}{0.08} & $\cellcolor{overlapcolor}{0.77}$ \asymci{0.04}{0.04} & $\cellcolor{bestcolor}{0.84}$ \asymci{0.04}{0.05} & $\cellcolor{overlapcolor}{0.74}$ \asymci{0.06}{0.06} \\
    \midrule
        assembly-v3 & $\cellcolor{bestcolor}{0.20}$ \asymci{0.43}{0.20} & $\cellcolor{overlapcolor}{0.00}$ \asymci{0.00}{0.00} & $\cellcolor{overlapcolor}{0.00}$ \asymci{0.00}{0.00} & $\cellcolor{overlapcolor}{0.00}$ \asymci{0.17}{0.00} \\
    basketball-v3 & $0.01$ \asymci{0.15}{0.01} & $0.69$ \asymci{0.17}{0.20} & $\cellcolor{overlapcolor}{0.97}$ \asymci{0.03}{0.06} & $\cellcolor{bestcolor}{0.97}$ \asymci{0.03}{0.11} \\
    button-press-v3 & $\cellcolor{bestcolor}{1.00}$ \asymci{0.00}{0.05} & $\cellcolor{overlapcolor}{0.96}$ \asymci{0.03}{0.18} & $\cellcolor{overlapcolor}{1.00}$ \asymci{0.00}{0.00} & $\cellcolor{overlapcolor}{1.00}$ \asymci{0.00}{0.01} \\
    coffee-pull-v3 & $\cellcolor{overlapcolor}{0.92}$ \asymci{0.06}{0.40} & $\cellcolor{overlapcolor}{0.94}$ \asymci{0.04}{0.08} & $\cellcolor{bestcolor}{0.97}$ \asymci{0.03}{0.07} & $0.81$ \asymci{0.08}{0.08} \\
    coffee-push-v3 & $\cellcolor{overlapcolor}{0.67}$ \asymci{0.18}{0.33} & $\cellcolor{overlapcolor}{0.70}$ \asymci{0.09}{0.10} & $\cellcolor{bestcolor}{0.80}$ \asymci{0.08}{0.05} & $\cellcolor{overlapcolor}{0.76}$ \asymci{0.12}{0.09} \\
    disassemble-v3 & $\cellcolor{overlapcolor}{0.14}$ \asymci{0.38}{0.14} & $\cellcolor{bestcolor}{0.44}$ \asymci{0.40}{0.40} & $0.00$ \asymci{0.01}{0.00} & $\cellcolor{overlapcolor}{0.00}$ \asymci{0.17}{0.00} \\
    hammer-v3 & $\cellcolor{bestcolor}{1.00}$ \asymci{0.00}{0.17} & $\cellcolor{overlapcolor}{0.99}$ \asymci{0.01}{0.16} & $\cellcolor{overlapcolor}{1.00}$ \asymci{0.00}{0.30} & $\cellcolor{overlapcolor}{1.00}$ \asymci{0.00}{0.17} \\
    hand-insert-v3 & $\cellcolor{overlapcolor}{0.71}$ \asymci{0.09}{0.10} & $\cellcolor{overlapcolor}{0.75}$ \asymci{0.18}{0.20} & $\cellcolor{bestcolor}{0.93}$ \asymci{0.07}{0.25} & $\cellcolor{overlapcolor}{0.66}$ \asymci{0.17}{0.12} \\
    handle-pull-v3 & $\cellcolor{overlapcolor}{0.94}$ \asymci{0.02}{0.11} & $\cellcolor{overlapcolor}{0.78}$ \asymci{0.14}{0.25} & $\cellcolor{bestcolor}{0.95}$ \asymci{0.05}{0.30} & $\cellcolor{overlapcolor}{0.54}$ \asymci{0.33}{0.42} \\
    lever-pull-v3 & $\cellcolor{overlapcolor}{0.72}$ \asymci{0.12}{0.17} & $\cellcolor{bestcolor}{0.76}$ \asymci{0.17}{0.16} & $\cellcolor{overlapcolor}{0.60}$ \asymci{0.17}{0.23} & $\cellcolor{overlapcolor}{0.53}$ \asymci{0.23}{0.15} \\
    peg-unplug-side-v3 & $\cellcolor{overlapcolor}{0.97}$ \asymci{0.03}{0.06} & $\cellcolor{overlapcolor}{0.88}$ \asymci{0.06}{0.07} & $\cellcolor{bestcolor}{0.99}$ \asymci{0.01}{0.05} & $\cellcolor{overlapcolor}{0.98}$ \asymci{0.02}{0.08} \\
    pick-out-of-hole-v3 & $\cellcolor{overlapcolor}{0.99}$ \asymci{0.01}{0.03} & $\cellcolor{overlapcolor}{0.96}$ \asymci{0.03}{0.08} & $\cellcolor{bestcolor}{1.00}$ \asymci{0.00}{0.03} & $\cellcolor{overlapcolor}{1.00}$ \asymci{0.00}{0.19} \\
    pick-place-v3 & $0.27$ \asymci{0.28}{0.21} & $0.63$ \asymci{0.16}{0.12} & $\cellcolor{bestcolor}{0.90}$ \asymci{0.07}{0.10} & $\cellcolor{overlapcolor}{0.74}$ \asymci{0.18}{0.37} \\
    pick-place-wall-v3 & $\cellcolor{overlapcolor}{0.00}$ \asymci{0.00}{0.00} & $\cellcolor{overlapcolor}{0.00}$ \asymci{0.02}{0.00} & $\cellcolor{overlapcolor}{0.00}$ \asymci{0.01}{0.00} & $\cellcolor{overlapcolor}{0.00}$ \asymci{0.00}{0.00} \\
    push-back-v3 & $\cellcolor{overlapcolor}{0.82}$ \asymci{0.18}{0.47} & $\cellcolor{overlapcolor}{0.97}$ \asymci{0.03}{0.05} & $\cellcolor{overlapcolor}{0.97}$ \asymci{0.03}{0.20} & $\cellcolor{bestcolor}{0.99}$ \asymci{0.01}{0.17} \\
    push-v3 & $0.62$ \asymci{0.26}{0.33} & $0.82$ \asymci{0.12}{0.15} & $\cellcolor{overlapcolor}{0.96}$ \asymci{0.04}{0.09} & $\cellcolor{bestcolor}{0.99}$ \asymci{0.01}{0.04} \\
    shelf-place-v3 & $\cellcolor{overlapcolor}{0.00}$ \asymci{0.17}{0.00} & $\cellcolor{bestcolor}{0.38}$ \asymci{0.45}{0.38} & $\cellcolor{overlapcolor}{0.22}$ \asymci{0.45}{0.22} & $\cellcolor{overlapcolor}{0.00}$ \asymci{0.12}{0.00} \\
    soccer-v3 & $0.42$ \asymci{0.17}{0.19} & $0.61$ \asymci{0.08}{0.07} & $\cellcolor{overlapcolor}{0.72}$ \asymci{0.13}{0.20} & $\cellcolor{bestcolor}{0.80}$ \asymci{0.08}{0.08} \\
    stick-pull-v3 & $\cellcolor{overlapcolor}{0.08}$ \asymci{0.36}{0.08} & $\cellcolor{overlapcolor}{0.53}$ \asymci{0.31}{0.40} & $\cellcolor{bestcolor}{0.79}$ \asymci{0.08}{0.40} & $\cellcolor{overlapcolor}{0.55}$ \asymci{0.27}{0.35} \\
    stick-push-v3 & $\cellcolor{overlapcolor}{0.09}$ \asymci{0.46}{0.04} & $\cellcolor{bestcolor}{0.73}$ \asymci{0.17}{0.28} & $\cellcolor{overlapcolor}{0.47}$ \asymci{0.29}{0.26} & $\cellcolor{overlapcolor}{0.12}$ \asymci{0.48}{0.12} \\

\\
\end{tabular}
    \caption{IQM success with 95\% bootstrapped confidence interval across the 20 hardest Metaworld tasks, comparing SEAR with various chunk sizes. The best result is highlighted, results that are non-significantly worse than the best are lightly highlighted. All policies except for SEAR-1 use a fixed receding horizon of five actions during evaluation. On average, SEAR-10 performs best.}
    \label{tab:metaworld-20-table}
\end{table}

\begin{table}[]
    \centering
\begin{tabular}{Sl | c c c c c c}
     & SEAR-10 & CQN-AS-5 & CQN-AS-10 & SimbaV2 \\
    \midrule
    Average & $\cellcolor{bestcolor}{0.84}$ \asymci{0.04}{0.05} & $0.61$ \asymci{0.04}{0.05} & $0.47$ \asymci{0.06}{0.06} & $0.62$ \asymci{0.05}{0.06} \\
    \midrule
        assembly-v3 & $0.00$ \asymci{0.00}{0.00} & $\cellcolor{overlapcolor}{0.88}$ \asymci{0.05}{0.03} & $0.60$ \asymci{0.20}{0.35} & $\cellcolor{bestcolor}{0.92}$ \asymci{0.08}{0.08} \\
    basketball-v3 & $\cellcolor{bestcolor}{0.97}$ \asymci{0.03}{0.06} & $0.25$ \asymci{0.03}{0.10} & $0.15$ \asymci{0.03}{0.07} & $0.27$ \asymci{0.36}{0.17} \\
    button-press-v3 & $\cellcolor{bestcolor}{1.00}$ \asymci{0.00}{0.00} & $0.43$ \asymci{0.42}{0.30} & $0.15$ \asymci{0.20}{0.15} & $0.95$ \asymci{0.03}{0.07} \\
    coffee-pull-v3 & $\cellcolor{bestcolor}{0.97}$ \asymci{0.03}{0.08} & $0.60$ \asymci{0.15}{0.13} & $0.55$ \asymci{0.03}{0.00} & $\cellcolor{overlapcolor}{0.88}$ \asymci{0.10}{0.20} \\
    coffee-push-v3 & $\cellcolor{bestcolor}{0.80}$ \asymci{0.08}{0.05} & $0.43$ \asymci{0.15}{0.08} & $0.37$ \asymci{0.18}{0.13} & $\cellcolor{overlapcolor}{0.38}$ \asymci{0.37}{0.22} \\
    disassemble-v3 & $0.00$ \asymci{0.01}{0.00} & $\cellcolor{bestcolor}{0.92}$ \asymci{0.07}{0.62} & $\cellcolor{overlapcolor}{0.35}$ \asymci{0.58}{0.35} & $\cellcolor{overlapcolor}{0.62}$ \asymci{0.08}{0.42} \\
    hammer-v3 & $\cellcolor{bestcolor}{1.00}$ \asymci{0.00}{0.30} & $\cellcolor{overlapcolor}{0.98}$ \asymci{0.02}{0.07} & $\cellcolor{overlapcolor}{1.00}$ \asymci{0.00}{0.03} & $\cellcolor{overlapcolor}{0.97}$ \asymci{0.03}{0.05} \\
    hand-insert-v3 & $\cellcolor{bestcolor}{0.93}$ \asymci{0.07}{0.24} & $\cellcolor{overlapcolor}{0.73}$ \asymci{0.10}{0.15} & $\cellcolor{overlapcolor}{0.72}$ \asymci{0.17}{0.55} & $\cellcolor{overlapcolor}{0.53}$ \asymci{0.32}{0.22} \\
    handle-pull-v3 & $\cellcolor{overlapcolor}{0.95}$ \asymci{0.05}{0.31} & $\cellcolor{bestcolor}{0.98}$ \asymci{0.02}{0.05} & $\cellcolor{overlapcolor}{0.98}$ \asymci{0.02}{0.03} & $0.80$ \asymci{0.08}{0.27} \\
    lever-pull-v3 & $\cellcolor{bestcolor}{0.60}$ \asymci{0.14}{0.20} & $\cellcolor{overlapcolor}{0.42}$ \asymci{0.08}{0.18} & $0.12$ \asymci{0.23}{0.07} & $\cellcolor{overlapcolor}{0.45}$ \asymci{0.17}{0.12} \\
    peg-unplug-side-v3 & $\cellcolor{bestcolor}{0.99}$ \asymci{0.01}{0.04} & $0.85$ \asymci{0.07}{0.08} & $\cellcolor{overlapcolor}{0.88}$ \asymci{0.12}{0.13} & $0.72$ \asymci{0.05}{0.22} \\
    pick-out-of-hole-v3 & $\cellcolor{bestcolor}{1.00}$ \asymci{0.00}{0.02} & $0.57$ \asymci{0.22}{0.12} & $0.62$ \asymci{0.05}{0.32} & $0.68$ \asymci{0.08}{0.17} \\
    pick-place-v3 & $\cellcolor{bestcolor}{0.90}$ \asymci{0.07}{0.10} & $0.58$ \asymci{0.18}{0.17} & $0.48$ \asymci{0.22}{0.40} & $0.00$ \asymci{0.17}{0.00} \\
    pick-place-wall-v3 & $\cellcolor{overlapcolor}{0.00}$ \asymci{0.01}{0.00} & $\cellcolor{overlapcolor}{0.12}$ \asymci{0.30}{0.12} & $\cellcolor{bestcolor}{0.18}$ \asymci{0.24}{0.18} & $\cellcolor{overlapcolor}{0.00}$ \asymci{0.33}{0.00} \\
    push-back-v3 & $\cellcolor{bestcolor}{0.97}$ \asymci{0.03}{0.19} & $\cellcolor{overlapcolor}{0.90}$ \asymci{0.08}{0.07} & $\cellcolor{overlapcolor}{0.97}$ \asymci{0.03}{0.65} & $\cellcolor{overlapcolor}{0.90}$ \asymci{0.08}{0.30} \\
    push-v3 & $\cellcolor{bestcolor}{0.96}$ \asymci{0.04}{0.09} & $0.70$ \asymci{0.05}{0.30} & $0.57$ \asymci{0.03}{0.05} & $0.65$ \asymci{0.16}{0.12} \\
    shelf-place-v3 & $\cellcolor{overlapcolor}{0.22}$ \asymci{0.43}{0.22} & $\cellcolor{bestcolor}{0.43}$ \asymci{0.40}{0.33} & $0.00$ \asymci{0.00}{0.00} & $\cellcolor{overlapcolor}{0.30}$ \asymci{0.37}{0.30} \\
    soccer-v3 & $\cellcolor{bestcolor}{0.72}$ \asymci{0.13}{0.19} & $\cellcolor{overlapcolor}{0.48}$ \asymci{0.08}{0.22} & $0.32$ \asymci{0.12}{0.13} & $\cellcolor{overlapcolor}{0.52}$ \asymci{0.16}{0.02} \\
    stick-pull-v3 & $\cellcolor{bestcolor}{0.79}$ \asymci{0.08}{0.39} & $\cellcolor{overlapcolor}{0.37}$ \asymci{0.05}{0.22} & $\cellcolor{overlapcolor}{0.32}$ \asymci{0.20}{0.29} & $\cellcolor{overlapcolor}{0.50}$ \asymci{0.25}{0.32} \\
    stick-push-v3 & $\cellcolor{overlapcolor}{0.47}$ \asymci{0.29}{0.27} & $\cellcolor{overlapcolor}{0.45}$ \asymci{0.12}{0.25} & $\cellcolor{bestcolor}{0.62}$ \asymci{0.18}{0.33} & $\cellcolor{overlapcolor}{0.25}$ \asymci{0.47}{0.17} \\
\\
\end{tabular}
    \caption{IQM success rate with 95\% bootstrapped confidence interval across the 20 hardest Metaworld tasks, comparing SEAR-10, the best performing SEAR variant, with baselines. The best result is highlighted, results that are non-significantly worse than the best are lightly highlighted. All policies except for SimbaV2 use a fixed receding horizon of five actions during evaluation. CQN-AS performs better with chunk size five (CQN-AS-5) than with chunk size ten (CQN-AS-10). Overall, SEAR-10 performs best.}
    \label{tab:metaworld-20-baselines-table}
\end{table}

\begin{table}[]
    \centering
\begin{tabular}{Sl | c c c c c c}
     & SEAR-1 & SEAR-5 & SEAR-10 & SEAR-20 & SimbaV2 \\
    \midrule
    Average & $\cellcolor{bestcolor}{1.00}$ \asymci{0.00}{0.00} & $0.99$ \asymci{0.00}{0.00} & $\cellcolor{overlapcolor}{1.00}$ \asymci{0.00}{0.00} & $\cellcolor{overlapcolor}{1.00}$ \asymci{0.00}{0.00} & $0.99$ \asymci{0.00}{0.01} \\
    \midrule
    bin-picking-v3 & $\cellcolor{bestcolor}{0.96}$ \asymci{0.04}{0.47} & $\cellcolor{overlapcolor}{0.89}$ \asymci{0.10}{0.27} & $\cellcolor{overlapcolor}{0.75}$ \asymci{0.20}{0.46} & $\cellcolor{overlapcolor}{0.53}$ \asymci{0.38}{0.40} & $\cellcolor{overlapcolor}{0.80}$ \asymci{0.18}{0.30} \\
    box-close-v3 & $\cellcolor{bestcolor}{1.00}$ \asymci{0.00}{0.03} & $0.83$ \asymci{0.05}{0.06} & $\cellcolor{overlapcolor}{0.95}$ \asymci{0.04}{0.10} & $\cellcolor{overlapcolor}{0.97}$ \asymci{0.03}{0.17} & $\cellcolor{overlapcolor}{0.92}$ \asymci{0.06}{0.08} \\
    button-press-topdown-v3 & $\cellcolor{bestcolor}{1.00}$ \asymci{0.00}{0.01} & $\cellcolor{overlapcolor}{1.00}$ \asymci{0.00}{0.03} & $\cellcolor{overlapcolor}{1.00}$ \asymci{0.00}{0.00} & $\cellcolor{overlapcolor}{1.00}$ \asymci{0.00}{0.00} & $\cellcolor{overlapcolor}{0.98}$ \asymci{0.02}{0.03} \\
    button-press-topdown-wall-v3 & $\cellcolor{bestcolor}{1.00}$ \asymci{0.00}{0.00} & $\cellcolor{overlapcolor}{1.00}$ \asymci{0.00}{0.01} & $\cellcolor{overlapcolor}{1.00}$ \asymci{0.00}{0.00} & $\cellcolor{overlapcolor}{1.00}$ \asymci{0.00}{0.00} & $\cellcolor{overlapcolor}{1.00}$ \asymci{0.00}{0.00} \\
    button-press-wall-v3 & $\cellcolor{bestcolor}{1.00}$ \asymci{0.00}{0.00} & $\cellcolor{overlapcolor}{1.00}$ \asymci{0.00}{0.00} & $\cellcolor{overlapcolor}{1.00}$ \asymci{0.00}{0.18} & $\cellcolor{overlapcolor}{1.00}$ \asymci{0.00}{0.01} & $\cellcolor{overlapcolor}{0.98}$ \asymci{0.02}{0.08} \\
    coffee-button-v3 & $\cellcolor{bestcolor}{1.00}$ \asymci{0.00}{0.03} & $\cellcolor{overlapcolor}{1.00}$ \asymci{0.00}{0.03} & $\cellcolor{overlapcolor}{1.00}$ \asymci{0.00}{0.01} & $\cellcolor{overlapcolor}{1.00}$ \asymci{0.00}{0.00} & $\cellcolor{overlapcolor}{1.00}$ \asymci{0.00}{0.00} \\
    dial-turn-v3 & $\cellcolor{bestcolor}{0.93}$ \asymci{0.05}{0.15} & $\cellcolor{overlapcolor}{0.81}$ \asymci{0.13}{0.21} & $\cellcolor{overlapcolor}{0.88}$ \asymci{0.10}{0.16} & $\cellcolor{overlapcolor}{0.88}$ \asymci{0.06}{0.18} & $\cellcolor{overlapcolor}{0.88}$ \asymci{0.10}{0.13} \\
    door-close-v3 & $\cellcolor{bestcolor}{1.00}$ \asymci{0.00}{0.00} & $\cellcolor{overlapcolor}{1.00}$ \asymci{0.00}{0.00} & $\cellcolor{overlapcolor}{1.00}$ \asymci{0.00}{0.01} & $\cellcolor{overlapcolor}{1.00}$ \asymci{0.00}{0.00} & $\cellcolor{overlapcolor}{1.00}$ \asymci{0.00}{0.00} \\
    door-lock-v3 & $\cellcolor{bestcolor}{1.00}$ \asymci{0.00}{0.00} & $\cellcolor{overlapcolor}{1.00}$ \asymci{0.00}{0.00} & $\cellcolor{overlapcolor}{1.00}$ \asymci{0.00}{0.02} & $\cellcolor{overlapcolor}{1.00}$ \asymci{0.00}{0.00} & $\cellcolor{overlapcolor}{1.00}$ \asymci{0.00}{0.00} \\
    door-open-v3 & $\cellcolor{bestcolor}{1.00}$ \asymci{0.00}{0.00} & $\cellcolor{overlapcolor}{0.98}$ \asymci{0.02}{0.09} & $\cellcolor{overlapcolor}{1.00}$ \asymci{0.00}{0.01} & $\cellcolor{overlapcolor}{1.00}$ \asymci{0.00}{0.01} & $\cellcolor{overlapcolor}{1.00}$ \asymci{0.00}{0.07} \\
    door-unlock-v3 & $\cellcolor{bestcolor}{1.00}$ \asymci{0.00}{0.00} & $\cellcolor{overlapcolor}{1.00}$ \asymci{0.00}{0.00} & $\cellcolor{overlapcolor}{1.00}$ \asymci{0.00}{0.01} & $\cellcolor{overlapcolor}{1.00}$ \asymci{0.00}{0.00} & $\cellcolor{overlapcolor}{1.00}$ \asymci{0.00}{0.03} \\
    drawer-close-v3 & $\cellcolor{bestcolor}{1.00}$ \asymci{0.00}{0.00} & $\cellcolor{overlapcolor}{1.00}$ \asymci{0.00}{0.00} & $\cellcolor{overlapcolor}{1.00}$ \asymci{0.00}{0.02} & $\cellcolor{overlapcolor}{1.00}$ \asymci{0.00}{0.01} & $\cellcolor{overlapcolor}{1.00}$ \asymci{0.00}{0.00} \\
    drawer-open-v3 & $0.22$ \asymci{0.50}{0.22} & $\cellcolor{overlapcolor}{0.83}$ \asymci{0.17}{0.44} & $0.27$ \asymci{0.45}{0.27} & $\cellcolor{overlapcolor}{0.94}$ \asymci{0.06}{0.46} & $\cellcolor{bestcolor}{1.00}$ \asymci{0.00}{0.00} \\
    faucet-close-v3 & $\cellcolor{bestcolor}{1.00}$ \asymci{0.00}{0.03} & $\cellcolor{overlapcolor}{1.00}$ \asymci{0.00}{0.04} & $\cellcolor{overlapcolor}{1.00}$ \asymci{0.00}{0.06} & $\cellcolor{overlapcolor}{1.00}$ \asymci{0.00}{0.00} & $\cellcolor{overlapcolor}{1.00}$ \asymci{0.00}{0.00} \\
    faucet-open-v3 & $\cellcolor{bestcolor}{1.00}$ \asymci{0.00}{0.03} & $\cellcolor{overlapcolor}{0.98}$ \asymci{0.02}{0.08} & $\cellcolor{overlapcolor}{1.00}$ \asymci{0.00}{0.03} & $\cellcolor{overlapcolor}{1.00}$ \asymci{0.00}{0.13} & $\cellcolor{overlapcolor}{1.00}$ \asymci{0.00}{0.07} \\
    handle-press-side-v3 & $\cellcolor{bestcolor}{1.00}$ \asymci{0.00}{0.00} & $\cellcolor{overlapcolor}{1.00}$ \asymci{0.00}{0.06} & $\cellcolor{overlapcolor}{1.00}$ \asymci{0.00}{0.00} & $\cellcolor{overlapcolor}{1.00}$ \asymci{0.00}{0.00} & $\cellcolor{overlapcolor}{1.00}$ \asymci{0.00}{0.00} \\
    handle-press-v3 & $\cellcolor{bestcolor}{1.00}$ \asymci{0.00}{0.03} & $\cellcolor{overlapcolor}{1.00}$ \asymci{0.00}{0.00} & $\cellcolor{overlapcolor}{1.00}$ \asymci{0.00}{0.00} & $\cellcolor{overlapcolor}{1.00}$ \asymci{0.00}{0.01} & $\cellcolor{overlapcolor}{1.00}$ \asymci{0.00}{0.03} \\
    handle-pull-side-v3 & $\cellcolor{bestcolor}{1.00}$ \asymci{0.00}{0.03} & $\cellcolor{overlapcolor}{0.92}$ \asymci{0.07}{0.13} & $\cellcolor{overlapcolor}{1.00}$ \asymci{0.00}{0.17} & $\cellcolor{overlapcolor}{1.00}$ \asymci{0.00}{0.03} & $\cellcolor{overlapcolor}{0.98}$ \asymci{0.02}{0.03} \\
    peg-insert-side-v3 & $\cellcolor{overlapcolor}{0.99}$ \asymci{0.01}{0.06} & $\cellcolor{overlapcolor}{0.97}$ \asymci{0.03}{0.04} & $\cellcolor{bestcolor}{1.00}$ \asymci{0.00}{0.01} & $\cellcolor{overlapcolor}{0.98}$ \asymci{0.02}{0.05} & $\cellcolor{overlapcolor}{0.90}$ \asymci{0.10}{0.52} \\
    plate-slide-back-side-v3 & $\cellcolor{bestcolor}{1.00}$ \asymci{0.00}{0.00} & $\cellcolor{overlapcolor}{1.00}$ \asymci{0.00}{0.00} & $\cellcolor{overlapcolor}{1.00}$ \asymci{0.00}{0.00} & $\cellcolor{overlapcolor}{1.00}$ \asymci{0.00}{0.00} & $\cellcolor{overlapcolor}{1.00}$ \asymci{0.00}{0.03} \\
    plate-slide-back-v3 & $\cellcolor{bestcolor}{1.00}$ \asymci{0.00}{0.00} & $\cellcolor{overlapcolor}{0.97}$ \asymci{0.03}{0.27} & $\cellcolor{overlapcolor}{1.00}$ \asymci{0.00}{0.00} & $\cellcolor{overlapcolor}{1.00}$ \asymci{0.00}{0.00} & $\cellcolor{overlapcolor}{1.00}$ \asymci{0.00}{0.00} \\
    plate-slide-side-v3 & $\cellcolor{overlapcolor}{0.90}$ \asymci{0.06}{0.07} & $\cellcolor{overlapcolor}{0.82}$ \asymci{0.08}{0.07} & $\cellcolor{overlapcolor}{0.82}$ \asymci{0.07}{0.08} & $\cellcolor{bestcolor}{0.95}$ \asymci{0.05}{0.12} & $\cellcolor{overlapcolor}{0.82}$ \asymci{0.10}{0.17} \\
    plate-slide-v3 & $\cellcolor{bestcolor}{1.00}$ \asymci{0.00}{0.00} & $\cellcolor{overlapcolor}{1.00}$ \asymci{0.00}{0.01} & $\cellcolor{overlapcolor}{1.00}$ \asymci{0.00}{0.00} & $\cellcolor{overlapcolor}{1.00}$ \asymci{0.00}{0.00} & $\cellcolor{overlapcolor}{1.00}$ \asymci{0.00}{0.00} \\
    push-wall-v3 & $\cellcolor{overlapcolor}{0.98}$ \asymci{0.02}{0.04} & $\cellcolor{overlapcolor}{0.95}$ \asymci{0.04}{0.12} & $\cellcolor{bestcolor}{1.00}$ \asymci{0.00}{0.01} & $\cellcolor{overlapcolor}{0.96}$ \asymci{0.04}{0.24} & $0.78$ \asymci{0.05}{0.10} \\
    reach-v3 & $\cellcolor{bestcolor}{1.00}$ \asymci{0.00}{0.01} & $\cellcolor{overlapcolor}{1.00}$ \asymci{0.00}{0.02} & $\cellcolor{overlapcolor}{1.00}$ \asymci{0.00}{0.00} & $\cellcolor{overlapcolor}{1.00}$ \asymci{0.00}{0.16} & $\cellcolor{overlapcolor}{0.95}$ \asymci{0.05}{0.05} \\
    reach-wall-v3 & $\cellcolor{bestcolor}{0.98}$ \asymci{0.02}{0.03} & $\cellcolor{overlapcolor}{0.98}$ \asymci{0.02}{0.04} & $\cellcolor{overlapcolor}{0.98}$ \asymci{0.02}{0.08} & $\cellcolor{overlapcolor}{0.96}$ \asymci{0.03}{0.20} & $\cellcolor{overlapcolor}{0.97}$ \asymci{0.03}{0.05} \\
    sweep-into-v3 & $\cellcolor{overlapcolor}{0.96}$ \asymci{0.04}{0.09} & $\cellcolor{overlapcolor}{0.96}$ \asymci{0.04}{0.06} & $\cellcolor{bestcolor}{1.00}$ \asymci{0.00}{0.05} & $\cellcolor{overlapcolor}{0.99}$ \asymci{0.01}{0.09} & $0.75$ \asymci{0.15}{0.17} \\
    sweep-v3 & $\cellcolor{overlapcolor}{0.97}$ \asymci{0.03}{0.10} & $\cellcolor{bestcolor}{1.00}$ \asymci{0.00}{0.05} & $\cellcolor{overlapcolor}{1.00}$ \asymci{0.00}{0.01} & $\cellcolor{overlapcolor}{1.00}$ \asymci{0.00}{0.00} & $\cellcolor{overlapcolor}{0.88}$ \asymci{0.08}{0.15} \\
    window-close-v3 & $\cellcolor{bestcolor}{1.00}$ \asymci{0.00}{0.00} & $\cellcolor{overlapcolor}{1.00}$ \asymci{0.00}{0.02} & $\cellcolor{overlapcolor}{1.00}$ \asymci{0.00}{0.00} & $\cellcolor{overlapcolor}{1.00}$ \asymci{0.00}{0.00} & $\cellcolor{overlapcolor}{1.00}$ \asymci{0.00}{0.00} \\
    window-open-v3 & $\cellcolor{bestcolor}{1.00}$ \asymci{0.00}{0.01} & $\cellcolor{overlapcolor}{0.99}$ \asymci{0.01}{0.03} & $\cellcolor{overlapcolor}{1.00}$ \asymci{0.00}{0.01} & $\cellcolor{overlapcolor}{1.00}$ \asymci{0.00}{0.00} & $\cellcolor{overlapcolor}{1.00}$ \asymci{0.00}{0.00} \\
    \\
\end{tabular}
    \caption{IQM success rate with 95\% bootstrapped confidence interval across the 30 easiest Metaworld tasks, comparing SEAR variants with a non-chunking SimbaV2 baseline. On nearly all tasks, the policies achieve close to a perfect success rate. We therefore exclude these environments from our other experiments.}
    \label{tab:metaworld-easy-table}
\end{table}

\clearpage
\section*{NeurIPS Paper Checklist}

\begin{enumerate}

\item {\bf Claims}
    \item[] Question: Do the main claims made in the abstract and introduction accurately reflect the paper's contributions and scope?
    \item[] Answer: \answerYes{}
    \item[] Justification: The claims made in the abstract and introduction are supported by the experimental results in Section~\ref{sec:results}. While actor state coverage is hard to quantify and changes throughout training, we quantify its effect on receding horizon effectiveness in Figure~\ref{fig:replan_result}. 

\item {\bf Limitations}
    \item[] Question: Does the paper discuss the limitations of the work performed by the authors?
    \item[] Answer: \answerYes{}
    \item[] Justification: The limitations of the method are discussed in Section~\ref{sec:conclusion}.

\item {\bf Theory assumptions and proofs}
    \item[] Question: For each theoretical result, does the paper provide the full set of assumptions and a complete (and correct) proof?
    \item[] Answer: \answerNA{}
    \item[] Justification: The paper does not present theoretical results.

    \item {\bf Experimental result reproducibility}
    \item[] Question: Does the paper fully disclose all the information needed to reproduce the main experimental results of the paper to the extent that it affects the main claims and/or conclusions of the paper (regardless of whether the code and data are provided or not)?
    \item[] Answer: \answerYes{}
    \item[] Justification: Implementations of the presented methods including source code, dependency versions and configuration files are provided at \url{https://alrhub.github.io/SEAR/}. All software used, especially the benchmarks are freely accessible.

\item {\bf Open access to data and code}
    \item[] Question: Does the paper provide open access to the data and code, with sufficient instructions to faithfully reproduce the main experimental results, as described in supplemental material?
    \item[] Answer: \answerYes{}
    \item[] Justification: Implementations of the presented methods including source code, dependency versions and configuration files are provided at \url{https://alrhub.github.io/SEAR/}. All software used, especially the benchmarks are freely accessible.

\item {\bf Experimental setting/details}
    \item[] Question: Does the paper specify all the training and test details (e.g., data splits, hyperparameters, how they were chosen, type of optimizer) necessary to understand the results?
    \item[] Answer: \answerYes{}
    \item[] Justification: The experimental setup is described in Section~\ref{sec:results}, with hyperparameters and further details provided in Appendix~\ref{app:exp_details}. Furthermore, the configuration files and implementations needed to reproduce our results are part of our code release (Appendix~\ref{app:implementation}).

\item {\bf Experiment statistical significance}
    \item[] Question: Does the paper report error bars suitably and correctly defined or other appropriate information about the statistical significance of the experiments?
    \item[] Answer: \answerYes{}
    \item[] Justification: We follow best practices and report IQM~\cite{agarwal2021deep} success with 95\% bootstrap confidence intervals computed over 500 samples.

\item {\bf Experiments compute resources}
    \item[] Question: For each experiment, does the paper provide sufficient information on the computer resources (type of compute workers, memory, time of execution) needed to reproduce the experiments?
    \item[] Answer: \answerYes{}
    \item[] Justification: We discuss the compute resources needed to reproduce our results in Appendix~\ref{app:compute}.
    
\item {\bf Code of ethics}
    \item[] Question: Does the research conducted in the paper conform, in every respect, with the NeurIPS Code of Ethics \url{https://neurips.cc/public/EthicsGuidelines}?
    \item[] Answer: \answerYes{}
    \item[] Justification: We use neither human subjects nor do we publish any datasets. SEAR has no societal impacts that sets it apart from other RL methods, e.g. possible replacement of jobs.

\item {\bf Broader impacts}
    \item[] Question: Does the paper discuss both potential positive societal impacts and negative societal impacts of the work performed?
    \item[] Answer: \answerNA{}
    \item[] Justification: Our research concerns itself with algorithmic improvements for RL which have no immediate impact on society.
    
\item {\bf Safeguards}
    \item[] Question: Does the paper describe safeguards that have been put in place for responsible release of data or models that have a high risk for misuse (e.g., pre-trained language models, image generators, or scraped datasets)?
    \item[] Answer: \answerNA{}
    \item[] Justification: We release no models or data that have a risk of misuse.

\item {\bf Licenses for existing assets}
    \item[] Question: Are the creators or original owners of assets (e.g., code, data, models), used in the paper, properly credited and are the license and terms of use explicitly mentioned and properly respected?
    \item[] Answer: \answerYes{}
    \item[] Justification: All assets used (benchmarks, algorithms, implementations) are properly cited, and their licenses are listed in Appendix~\ref{app:licenses}.

\item {\bf New assets}
    \item[] Question: Are new assets introduced in the paper well documented and is the documentation provided alongside the assets?
    \item[] Answer: \answerYes{}
    \item[] Justification: The code released contains suitable documentation.

\item {\bf Crowdsourcing and research with human subjects}
    \item[] Question: For crowdsourcing experiments and research with human subjects, does the paper include the full text of instructions given to participants and screenshots, if applicable, as well as details about compensation (if any)? 
    \item[] Answer: \answerNA{}
    \item[] Justification: We neither use crowdsourcing nor human subjects in this work.

\item {\bf Institutional review board (IRB) approvals or equivalent for research with human subjects}
    \item[] Question: Does the paper describe potential risks incurred by study participants, whether such risks were disclosed to the subjects, and whether Institutional Review Board (IRB) approvals (or an equivalent approval/review based on the requirements of your country or institution) were obtained?
    \item[] Answer: \answerNA{}
    \item[] Justification: We neither use crowdsourcing nor human subjects in this work.

\item {\bf Declaration of LLM usage}
    \item[] Question: Does the paper describe the usage of LLMs if it is an important, original, or non-standard component of the core methods in this research? Note that if the LLM is used only for writing, editing, or formatting purposes and does \emph{not} impact the core methodology, scientific rigor, or originality of the research, declaration is not required.
    \item[] Answer: \answerNA{}
    \item[] Justification: Our method does not involve LLMs.

\end{enumerate}

\end{document}